\documentclass[review, 12pt]{elsarticle}

\usepackage{amssymb}

\usepackage[T1]{fontenc}
\usepackage{amsmath,amssymb,amsfonts}
\usepackage{graphicx}
\usepackage{textcomp}
\usepackage{xcolor}
\usepackage{multirow}

\usepackage{url}
\usepackage{svg}
\usepackage{subcaption}
\usepackage[shortlabels]{enumitem}
\usepackage{siunitx}
\usepackage{adjustbox}

\def\BibTeX{{\rm B\kern-.05em{\sc i\kern-.025em b}\kern-.08em
    T\kern-.1667em\lower.7ex\hbox{E}\kern-.125emX}}

\journal{Knowledge-Based Systems}
\begin{document}

\begin{frontmatter}



\cortext[cor]{Corresponding author}

\title{A Natural Gas Consumption Forecasting System for Continual Learning Scenarios based on Hoeffding Trees with Change Point Detection Mechanism}
\author[label1]{Radek Svoboda\corref{cor}}
\ead{radek.svoboda@vsb.cz}
\author[label2]{Sebasti\'an Basterrech}
\ead{sebbas@dtu.dk}
\author[label3]{Jędrzej Kozal}
\ead{jedrzej.kozal@pwr.edu.pl}
\author[label1]{Jan Platoš}
\ead{jan.platos@vsb.cz}
\author[label3]{Michał Woźniak}
\ead{michal.wozniak@pwr.edu.pl}
\affiliation[label1]{organization={Faculty of Electrical Engineering  and Computer Science, V\v{S}B--Technical University of Ostrava},
addressline={17. listopadu 2172/15}, 
            city={Ostrava},
            postcode={70833}, 
            country={Czechia}}
\affiliation[label2]{organization={Department of Applied Mathematics and Computer Science, Technical University of Denmark},
             city={Kongens Lyngby},
             country={Denmark}
}
\affiliation[label3]{organization={Department of Systems and Computer Networks,  University of Science and Technology},
             city={Wroclaw},
             country={Poland}
}


\begin{abstract}

Forecasting natural gas consumption, considering seasonality and trends, is crucial in planning its supply and consumption and optimizing the cost of obtaining it, mainly by industrial entities. However, in times of threats to its supply, it is also a critical element that guarantees the supply of this raw material to meet individual consumers' needs, ensuring society's energy security. This article introduces a novel multistep forecasting of natural gas consumption with change point detection integration for model collection selection with continual learning capabilities using data stream processing. The performance of the forecasting models based on the proposed approach is evaluated in a complex real-world use case of natural gas consumption forecasting. Furthermore, the methodology generability was verified in an electricity load forecasting task. We employed Hoeffding tree predictors as forecasting models and the Pruned Exact Linear Time (PELT) algorithm for the change point detection procedure. The change point detection integration enables the selection of a different model collection for successive time frames. Thus, three model collection selection procedures are defined and evaluated for forecasting scenarios with various densities of detected change points. These models were compared with change point agnostic baseline approaches and deep learning models. Our experiments show that the proposed approach provides superior results to deep learning models for both datasets and that fewer change points result in a lower forecasting error regardless of the model collection selection procedure employed.
\end{abstract}

\begin{keyword}
incremental learning \sep time series forecasting \sep multivariate time series \sep change point detection \sep machine learning \sep data stream processing

\end{keyword}

\end{frontmatter}


\newcommand{\Process}{\mathcal{P}}
\newcommand{\z}{Z}
\newcommand{\Error}{E}


\section{Introduction}
As a result of ongoing trends, the energy industry is experiencing a growing demand for consumption forecasting tools. In the past, these tools focused mainly on electricity load forecasting, as they presented complex and costly storage challenges. Consequently, the development of similar tools for other areas was postponed. However, the importance of natural gas is now on the rise, driven by the increasing environmental initiatives undertaken by governments and private companies. Despite this, effective logistics and delivery planning in the natural gas sector presents significant challenges.

A primary driver for energy consumption forecasting stems from the prevalent use of take-or-pay contracts and clauses in commodity trading \citep{creti2004long, medina1986take}. These contractual agreements require customers to pay for the specified quantity, regardless of actual usage. The financial consequences of such agreements underscore the importance of improving the quality of the forecasting models. The vital landscape of energy consumption shows a need for more effective forecasting models with continual adaptation capabilities. This adaptability ensures that the models remain relevant and effective in capturing the nuances of changing consumption patterns, market dynamics, and external influences. This responsiveness to new data improves the overall reliability of the system. Creating a more resilient framework capable of addressing the challenges inherent in the energy sector. Therefore, recognizing the pivotal role of continual model adaptation is crucial for stakeholders navigating the intricacies of energy consumption forecasting in the context of take-or-pay contracts.


Change point detection techniques are useful for understanding and predicting patterns in time series forecasting. 
A change point event refers to a specific moment in a sequence where the underlying behavior or trend shifts abruptly. 
Various factors can cause these shifts, such as policy changes, economic events, technological advancements, or changes in consumer behavior. 
Identifying change points is crucial to developing an accurate and robust forecasting model. Change points can significantly impact the underlying distribution of time series. In the particular case of energy consumption, it can cause modifications in the near-term energy demand.
Furthermore, understanding change points and seasonality in time series forecasting enables energy industry stakeholders to make more informed decisions. 
%
%
By identifying abrupt changes in consumption behavior, energy providers can anticipate shifts in energy demand, adjust their strategies accordingly, and improve energy grid management.

Recently, Continual Learning (CL) holds great importance in 
time series forecasting. 
With technological advances and the availability of large-scale datasets, CL allows forecasting methods to incorporate advanced approaches and continuously integrate new information, enhancing their forecasting capabilities.
The ability to adapt and learn from new data and methodologies ensures that forecasting tools remain reliable and effective in meeting the evolving demands of the energy industry.
%
%
By continually updating and refining predictive models in the context of CL, predictive tools can capture distribution changes in input patterns and provide different weights to the latest data, enabling more accurate predictions of energy demand.

%
The main contributions of the work are listed below.
\begin{itemize}
    \item[(i)] We introduce a general framework for multistep forecasting of multivariate time series as CL task.
    The innovative framework is composed of several hierarchical independent modules, such as dataset preprocessing, change point detection, model collection setup, model collection selection schema, and model training, which are fully described in the paper. The modular architecture enhances flexibility and makes it easy to exchange specific modules if the characteristics of the problem change.
    %

    \item[(ii)] We study the performance of the framework in a wide range of scenarios. We investigate several approaches for forecasting natural gas consumption and electricity load, using the integration of the detection of change points with the selection approach of model collection (with or without an error feedback loop).
    
      %
    %
    \item[(iii)] We present a sensitive analysis of the integration of the PELT algorithm into the CL system. We also discuss the threat to validity of our research. 
    The experimental analysis includes a comparison with two baseline approaches that do not consider change point detection and also with deep learning forecasting models.
    Furthermore, we study the impact of number of detected change points on the overall predictive performance.
    The evaluation was carried out on recently collected real-world dataset from the energy distribution domain. 
    We evaluated the developed system using three standard metrics (MSE, MAE and SMAPE) and Diebold-Mariano test to better capture the time series behavior. 
\end{itemize}


%
\section{Related works}
\label{sec:rel_works}
This section focuses on four pivotal topics central to our research. Our primary emphasis lies in forecasting natural gas consumption, given its vital role in energy planning and trading. Additionally, we delve into the literature on the CL paradigm in energy consumption forecasting. The Hoeffding Decision Tree is explored for its algorithmic relevance, and we review the literature on Change Point Detection methods for identifying abrupt shifts in the time series data. By synthesizing insights from these areas, our study aims to build a solid foundation for improving the predictive capabilities of natural gas consumption models.


\subsection{Natural gas consumption forecasting}
Statistical methods have gained popularity in forecasting gas consumption during the late 90s~\citep{Balestra1966}.
A nonlinear regression model was applied to model the gas consumption behavior of individual and small commercial customers~\citep{Vondracek2008}.
Due to the success of Neural Networks (NNs) for solving machine learning problems, several works also applied their potential. Oil market data and weather information were studied to create a forecasting model employing NNs for Belgian gas consumption~\citep{Suykens1996}.
Other families of NNs have also been studied, a two-stage model that combined feedforward and functional link NNs~\citep{Khotanzad2000}, while a hybrid of traditional networks and fuzzy NNs was developed~\citep{Viet2005}. 
Hourly sampled data was analyzed using linear methods, support vector machines, and forward NNs~\citep{Soldo2014}. The authors showed how different time windows of the day present different consumption trends.
Another learning application was over the gas consumption in Turkish regions~\citep{Taspinar2013,Boran2015}.
The SARIMAX model, multilayer perceptron ANN, and RBF models were evaluated for the prediction of daily gas consumption~\citep{Taspinar2013}.
The impact of including exogenous variables in specific forecasting models for the gas consumption was also investigated, especially the relevance of the air temperature feature~\citep{Taspinar2013}.
%
%
%

Multilayer feedforward neural network also showed to be effective as forecasting method of hourly gas consumption when the input layer has included features with weather data information~\citep{Szoplik2015}.
However, the relationship between external variables and gas consumption was not always evident and direct.
Several studies considered the thermal memory capacity of the buildings. For example, air temperature and solar radiation may have some delays in the impacts of gas consumption~\citep{Soldo2014,rahman2017}.
Forecasting based on the LSTM model and evolutionary algorithms for hyperparameter optimization were also studied~\citep{Su2019}.
%
%
%
In~\citep{WEI2023205133, WEI2024122087} hybrid LSTM-based models were proposed for transfer learning scenarios, the hybrid frameworks focus on preventing the negative transfer phenomenon using novel transfer domain selection algorithms.

Although NNs have been widely applied to solve the forecasting problem, other tools have also reported good results. 
Support vector machines were also employed with structure calibration~\citep{BAI2016}. 
According to their results, kernel-based methods outperformed classic backpropagation models in predicting daily gas consumption. 
Two reviews of the literature on the area of natural gas forecasting are available in~\citep{SOLDO2012,Tamba2018}.
%
%
%
%

\subsection{Continual Learning}
Despite recent advances in the machine learning domain and the large number of applications, most learning tools are designed to model stationary learning data. 
The CL paradigm has been introduced to overcome some classic machine learning limitations.
A CL approach defines an algorithm that can learn in incremental mode, which is suitable for not stationary data modeling~\citep{Chen:2018}. 
The greatest challenge in such a setting is overcoming catastrophic forgetting~\citep{FRENCH1999128} -- a tendency for NNs to quickly forget what they have learned from previously seen data, especially when those data are no longer available.
The optimal solution for the CL problem was shown to be NP-hard~\citep{Knoblauch2020}, and many heuristics have been proposed so far. 
Most of the current CL methods are based on rehearsal~\citep{PARISI201954}. 
In such methods, historical samples are stored in a small buffer and could be used to prevent forgetting when training with new data. Rehearsal algorithms can be quickly adapted to the online setting by utilization of reservoir sampling~\citep{DBLP:journals/corr/abs-1902-10486}. In~\citep{NEURIPS2019_15825aee}, instead of randomly selecting the learning examples from the rehearsal buffer when learning a new task, samples were selected based on the loss change when performing network updates. 
The authors of~\citep{caccia2022new} observed a shift of old data representations in the network feature space during training with new learning examples. They also introduced a method to correct this problem. 
The work of~\citep{NEURIPS2022_5ebbbac6} provided theoretical insight into the risk of overfitting when using a small memory buffer.
Dark experience replay~\citep{NEURIPS2020_b704ea2c} extended the standard rehearsal to save not only historical samples, but also logits with predictions that are later used for knowledge distillation. This method was later improved~\citep{9891836} by introducing bias correction, adding pre-training for future tasks, and updating existing logits in memory.
Other approaches to CL can include the inclusion of additional regularization terms~\citep{doi:10.1073/pnas.1611835114,si}, or the extension of the neural network architecture for new tasks~\citep{DBLP:journals/corr/RusuRDSKKPH16,ermis2022memory}.
The area is advancing fast, and CL was also previously applied to solve time series problems. In~\citep{9679108} a new method was introduced for multi-sensor time series based on task-specific generative models and classifiers. The authors of~\citep{10.1145/3517745.3563033} proposed a new method based on Variational Auto Encoders and Dilated Convolutional NNs aimed at multivariate time series anomaly detection problems. The work of~\citep{COSSU2021607} studied CL with recurrent NNs by empirically evaluating existing algorithms and introducing new benchmarks for CL with sequential data.

\section{Proposed methodology} \label{sec:methodology}
%
%






Existing research in energy consumption forecasting is predominantly focused on static or non-adaptive models, overlooking the importance of continual adaptation \citep{TONG2023106005, HUSSAIN2023101, en15134880}. Our methodology addresses this gap by emphasizing the challenges and methodologies associated with the continual improvements of forecasting models in the context of energy consumption. Moreover, we emphasize the integration of change point detection methods. Recognizing that sudden changes throughout the year characterize energy consumption, we highlight the critical role of change point detection in enhancing the adaptability of forecasting models. By incorporating change point detection methods, our approach goes beyond traditional static models, allowing the identification of significant shifts in consumption patterns.

\begin{figure*}[ht!]
  \centering
    \includegraphics[width=1.0\textwidth]{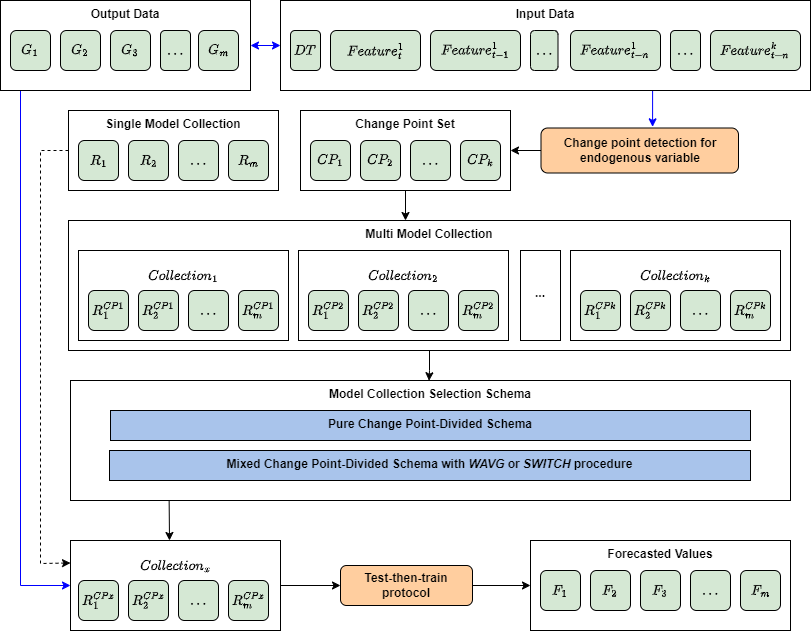}
  \caption{The proposed multistep forecasting pipeline with CL support.}
  \label{fig:forecast_pipeline}
\end{figure*}

A general methodology for multistep forecasting of multivariate time series with CL capabilities employing time series in the form of data stream processing is depicted in Fig.~\ref{fig:forecast_pipeline}.

%
%
The methodology consists of a modular schema which forms the following pipeline:

\noindent\textbf{Change point detection procedure.} This step is optional and is used in the Multi Collection scenario. This step aims to divide the time series into several segments. Each segment should have different statistical properties (e.g., trend direction, variance, or mean). Thus, each segment will use a different model collection for forecasting. Diving the time series into segments simplifies the forecasting task, as each model collection should be trained on data with similar properties. Change points can be set manually by knowledge regarding the domain (e.g., based on knowledge about the heating season in the natural gas consumption case) or by a change point detection algorithm such as PELT or Binary segmentation. The detected change point locations in a chosen period (e.g., the first year of the data) are then re-used in subsequent periods.

\noindent\textbf{Model collection setup.} The defined approaches may utilize one or more forecasting models. The forecasting model, in general, is a mathematical representation designed to make predictions about future values in a short, medium, or long-term time horizon based on historical data and patterns, for example, to forecast $n$ hours ahead of gas consumption or $m$ days ahead of electricity load. The first is a Single Model Collection approach, which does not divide the data into segments but uses only one model collection for the whole dataset. The model collection is an ordered set of partial models that provides forecasts for the defined forecast horizon (for the number of steps after the forecast origin). The number of models ($m$) in the collection is equal to the length of the forecast horizon. The reason for using direct forecasting with multiple models instead of the cumulative approach, which reuses the forecasts from the previous timestep in predicting the next value, is to prevent the accumulation of errors through the forecast horizon. The second approach uses Multi Model Collections. The number of collections ($k$) depends on the number of segments the time series was divided into. Thus, each segment has its own assigned model collection, i.e., if we divide the time series into $k$ segments, $k$ model collections will be used, one for each segment.
The rationale behind this approach is that there should be a benefit in having different strategies for distinct parts of the time series, as each segment might have different properties. The number of partial forecasting models is the same as in the Single Model Collection approach. The specific model used is a decision made while applying the methodology. The model has to support incremental learning, so models such as Hoeffding Tree or Stochastic Gradient Tree can be used for this task.

\noindent\textbf{Model collection selection procedure.} After the model collections are set, the only step left before the training phase is a logic for model collection selection. The simplest case is a Single Model Collection approach as it uses only one collection, and thus, this step could be skipped. There are two approaches for the collection selection for Multi-Model Collections, also called model aggregation schemes. In simple terms, the model aggregation scheme is the plan for deciding which set of prediction models to use based on the changes observed in the time series data. The first one, the Pure Change Point-Divided schema, selects the model collection based on the current time series segment (the segments are divided by change points from the \textit{Change point detection} step). When a change point occurs, the current model collection is exchanged for the subsequent one. Added the sentence laterThe rationale behind this schema is that when the time-series distribution shifts, switching to a different group of forecasting models is beneficial. The second approach, called Mixed Change Point-Divided schema, is based on the hypothesis that the change points do not occur at the same location in every period, but the location varies in some close subsegments. Thus, this approach works with a boundary of $b$ steps, which defines a segment of length $2b$ ($b$ step prior to the change point and $b$ steps after it). Both model collections (before and after the change point) are used inside this boundary, the first procedure uses weighted average ensembling (abbr. WAVG) and the second one is built upon alternative model switching (abbr. SWITCH). They are trained on the new instances and produce forecasts.

   For both model collections, a forecast error measure $\Error$ is computed using an arbitrarily selected metric (e.g., MAE, MSE, etc.). 
    Error calculation is used at timestep $t+1$ for the final forecast construction, and this process can be performed in one of two ways:
    \begin{enumerate}
        \item \textbf{WAVG procedure:} 
        Let $R_{WA}$ be a collection of $m$ forecasting models generated by the WAVG procedure \big($R^{(t)}_{WA} = (r^{(t)}_1, r^{(t)}_2, ..., r^{(t)}_m)$\big).
        Furthermore, we denote by $C_1$ and $C_2$ two disjoint collections of partial models $R^{(t)}_{C_1}$ and $R^{(t)}_{C_2}$ before and after a specific change point.
        The forecast value of model $r_i$ at timestep $t$ ($f_i^{(t)}$) is calculated as a weighted average of the forecasts produced by the following two model collections $R^{(t)}_{C_1}$ and $R^{(t)}_{C_2}$:
\begin{equation}
\label{eq:wavg}
    f^{(t)}_{i} = \frac{w_{C_1} f^{(t)}_{i,C_1} + w_{C_2} f^{(t)}_{i,C_2}}{w_{C_1} + w_{C_2}},
\end{equation}
where the  weights $w_{C_1}$ and $w_{C_2}$ are defined according to the previous performances (errors $E^{(t-1)}_{C_1}$ and $E^{(t-1)}_{C_2}$):

%
%
\begin{align} 
w_{C_1} &= 1 - \frac{\Error^{(t-1)}_{C_1}}{\Error^{(t-1)}_{C_1} + \Error^{(t-1)}_{C_2}}\nonumber,\\
w_{C_2} &= 1 - \frac{\Error^{(t-1)}_{C_2}}{\Error^{(t-1)}_{C_1} + \Error^{(t-1)}_{C_2}}.\nonumber
        \end{align}

       \item \textbf{SWITCH procedure:} Let $R_{SW}$ be a collection of $m$ forecasting models generated by the SWITCH procedure \big($R^{(t)}_{SW} = (r^{(t)}_1, r^{(t)}_2, ..., r^{(t)}_m)$\big). The forecast at the timestep $t$ is also based on the forecasts produced by the two model collections, as in the WAVG procedure case. However, in this procedure, only the forecast at the timestep $t$ by the model collection with lower $\Error$ at the timestep $t-1$ is used. Thus, the models switch during the boundary period. The forecast value of model $r_i$ ($f_i^{(t)}$) at the timestep $t$ is produced by the partial model of the selected collection ${C_1}$ or ${C_2}$:
       
        \begin{equation} 
        f^{(t)}_{i} = \begin{cases} f^{(t)}_{i,C_1}, \:\: \text{if  } \Error^{(t-1)}_{C_1} < \Error^{(t-1)}_{C_2}, \\ f^{(t)}_{i,C_2}, \:\: \text{otherwise.} \end{cases}
        \end{equation}
        
    \end{enumerate}



In both cases, these procedures help to decide how to select different prediction models based on their past performance around detected change points in the data. WAVG combines predictions from various models with weighted averages, giving more influence to historically accurate models. On the other hand, SWITCH selects predictions from one model based on its past performance, adapting to changing circumstances in the data.
After the boundary sub-segment ends, only a single model collection is used, and its choice is the same as in the Pure Change Point-Divided schema case.

\noindent\textbf{Model training.} When the model collections are set up, and the selection schema is selected, the model collections could be used in a training phase. The methodology is meant for data stream incremental learning; thus, the models are trained after each data instance. Before training, the models produce the forecasted values, and error metrics are computed and used if the Mixed Change Point-Divided schema is used.
%

\section{Algorithms utilized in the methdology application}
Algorithms serve as the backbone of any methodology. In this section, we provide an overview of the key algorithms employed within our methodology. Understanding their roles and functionalities is paramount to understanding the inner workings of our approach and its applicability in real-world scenarios.

\subsection{The Hoeffding decision tree}
Standard tree-based classification or regression algorithms are not well suited for online learning~\citep{Breiman1984ClassificationAR,Quinlan1986,10.5555/583200}. They learn by splitting the data into separate manifolds that contain either homogenous samples from a single class or low target variable variance. For this reason, updating the tree's structure is not as easy as in the case of NNs. Domingos and Hulten introduced a new method that allows tree structure modification~\citep{10.1145/347090.347107}. 
They recursively find the best attribute to split data in each node using only a small amount of training data. To calculate the number of samples in each node, that is necessary they employ Hoeffding bound in the form of:

\begin{equation}
    P( \bar{X} - E[X] \leq \sqrt{\frac{R^2 ln(1 / \delta)}{2n}} ) = 1 - \delta
\end{equation}
\noindent where $X$ is a random variable, $R$ is its range, and $n$ is the number of observations. This bound allows for obtaining an attribute with a high probability of providing a better split than the second-best attribute.
To limit the size of the decision tree, authors also introduce a memory limit that, when reached, triggers a tree pruning mechanism based on error reduction provided by a given tree leaf.
Domingos and Hulten also showed that trees optimized using an online schema are asymptotically similar to trees learned offline using the whole data.
A more detailed description can be found in \citep{10.1145/347090.347107}.

Another tree variation introduced fast splits with non-zero information gain~\citep{10.1145/3219819.3220005}.
The forecasting capacity of Streaming Random Forest with Hoeffding trees as base models was also investigated~\citep{randomforest}.
%
%
%
A parallel version of a Hoeffding tree was proposed that computes information gain on multiple computational nodes, decreasing the overall computing time~\citep{10.1007/978-3-319-00551-5_4}.
%
%

\subsection{Change point detection}
%
%
Let $(X_t)_{t\geq 0}$ be a time series of  independent and identically distributed (i.i.d.) observations. 
%
%
In a non-stationary environment, it may occur that the probability distribution of $X$ changes over time, then there exists a point $t\in \{0,1,...\}$ such that the underlying distribution of $\{\ldots,X_{t-2},X_{t-1},X_t\}$ is different from the distribution of $\{X_{t+\Delta},X_{t+\Delta+1}, \ldots\}$, for a $\Delta>0$.
A change point detection task is based on minimizing a cost function that should evaluate the homogeneity of the found segments, and they are usually linked to parametric probabilistic models (e.g., using maximum likelihood method) or non-parametric approaches (as rank-based algorithms)~\citep{Lung-Yut-Fong:2015}. Another critical component is a search method that could return optimal or approximate solutions~\citep{Truong:2020}. 
Some algorithms also employ penalty terms, which are responsible for minimizing overfitting~\citep{Haynes:2014}. Older methods that require knowing the number of change points may not be practical in real-world applications.
Multiple approaches have been proposed to solve the problem of detecting change points~\citep{Aminikhanghahi:2017}.
However, the most well-known method seems to be the
Pruned Exact Linear Time (PELT)~\citep{Killick:2012} algorithm. It could be seen as an improvement of the Optimal Partitioning method~\citep{Jackson:2005}.

%
%
%
%
It minimizes the cost function:
\begin{equation}
   \sum_{i=1}^{M+1}[C(y_{\tau_{i-1}+1})+\beta],
\end{equation}
\noindent where $\tau$ is a change point and $M$ is the total number of change points, $\beta$ is a penalization parameter.

The partitioning procedure uses a pruning method that improves computational efficiency by reducing the number of change points managed in a single iteration.
There is a plethora of other change point detectors that have been proposed so far, such as At Most Single Changepoint (AMOC), Binary Segmentation algorithm~\citep{Scott:1974}, or Bayesian Online Changepoint Detection~\citep{Adams:2007}.

Interest in these methods is not waning, and attempts are being made to develop change point detection methods for increasingly complex problems, including taking into account not only the non-stationarity of time series but also the fact that nowadays, more and more tasks could be multidimensional~\citep{Aminikhanghahi:2017}. An interesting proposal is the WATCH algorithm that allows change point detection for multidimensional time series~\citep{Faber:2021}. The solution is based on analyzing distances between distributions using the Wasserstein metric.
Also, interesting proposals using specific penalty functions may be found. Several works provide a recipe for estimating an interval with a useful penalty value -- see, for example, the CROPS algorithm~\citep{Haynes:2014} or the ALPIN scheme for optimal penalty setting~\citep{Truong:2020}.

%
%
%
%
%
%

\section{Methodological insights and validity threats}
Identifying and mitigating validity threats is crucial for reliable predictive models. Our research uses a unique real-world dataset in a CL context, offering a fresh perspective but complicating comparisons with other studies. This scarcity of previous research adds originality, but presents challenges in benchmarking our results. Therefore, we contrast our methodology and results with other methods in terms of model adaptability, feature engineering, analysis approach, and model accuracy.

\subsection{Threats to validity}
In the realm of empirical research focused on forecasting in the energy domain, the identification and mitigation of validity threats play a crucial role in ensuring the reliability and applicability of predictive models. Our goal is to create accurate forecasts capable of informing real-world decision-making, and it becomes imperative to recognize potential challenges that could compromise the validity of our results. 
We consider the following threats to be the most severe in relation to our research.

\subsubsection{Data selection bias}
\noindent\textbf{Connection to the research.} In our context, data selection bias could manifest itself if the data set predominantly represents certain seasons, leading to skewed patterns in the time series. For example, if the data set is predominantly from colder seasons, the model may not generalize well to warmer seasons, affecting the reliability of the forecasting system.

\noindent\textbf{Mitigation strategy.} To address data selection bias, we ensured that the seasons are equally represented; the same applies to any medium- or long-term trend, using a dataset that contains data from 2013 to 2020. Besides the natural gas consumption data, the experiments were also carried out using electricity load data to ensure that the proposed methodology can be adopted in different tasks and is not limited to a single domain.

\subsubsection{Methodological bias}
\noindent\textbf{Connection to the research.} Both machine learning and change point detection algorithms may be sensitive to hyperparameter settings which forms potential source of methodological bias because it may severely affect the results obtained. There may be a risk of selection bias present in the evaluation metrics and selection of the train-test sets in the experiment design phase.

\noindent\textbf{Mitigation strategy.} Sensitivity analyses evaluated the impact of different methodological choices on model performance. Various configurations for the number of change points and multiple approaches were tested. To avoid cherry-picking metrics that favor the model, multiple widely used error metrics (MAE, MSE, SMAPE) were reported. An interleaved test-then-train protocol was used to assess generalization performance, ensuring the model was not overly tailored to specific data subsets.

\subsubsection{Confounding features threat}
\noindent\textbf{Connection to the research.} Confounding feature threats arise when external factors (exogenous variables) that could affect predictions are not considered in the methodological design. External factors, such as weather, temperature, forecast for the next day, or date-time features can influence natural gas consumption or electricity load.

\noindent\textbf{Mitigation strategy.} We have conducted extensive research in this domain in our work \citep{SVOBODA2021119430} to identify potential external factors that influence the endogenous variable. Relevant exogenous variables were incorporated into the model.

\subsection{Methodology and results in context}
As described in Section~\ref{sec:rel_works}, many existing studies are based on conventional datasets \citep{Taspinar2013, TASCIKARAOGLU2014243}. In contrast, the proposed research uses a unique and underexplored real-world dataset \citep{SVOBODA2021119430} in a CL context, introducing a new perspective to the field of energy consumption forecasting. The scarcity of prior research on this specific dataset adds to the originality of our work, but also complicates the comparison of forecast accuracy with related studies. Therefore, we contrasted our methodology and results with other methods in terms of model adaptability, feature engineering, approach to analysis, and model accuracy.

\subsubsection{Forecasting model adaptability}
\noindent\textbf{Research context.} The prevailing trend in related research within the realm of energy consumption forecasting predominantly focuses on static models with limited adaptation capabilities \citep{Taspinar2013}. Moreover, these models are typically used for single-point forecasts; thus, they provide forecasts as scalars for a single point in time \citep{bento9854716}.

\noindent\textbf{Proposed methods contrasts.} Unlike related approaches that use static models for single-point forecasts, our research adopts a different strategy. We apply CL principles to treat natural gas consumption or electricity load as a continuous data stream, allowing our models to learn and adapt in real-time. Our method focuses on vector forecasts, predicting an entire forecast horizon rather than just one time unit ahead. Specifically, we target a 24-hour forecast horizon, which is standard in this domain.

\subsubsection{Feature selection possibilities}
\noindent\textbf{Research context.} Related research often uses statistical methods and neural networks for natural gas consumption forecasting. Statistical models like ARIMA \citep{HOSOVSKY2021101955} rely on univariate time series, limiting their ability to capture complex multivariate relationships. Neural networks, while powerful, struggle to adapt to changing input data distributions, require longer input sequences, offer limited feature engineering, and generally depend on offline learning with the entire available dataset \citep{woo2023deep}. The conventional practice in the energy domain is to use a single forecasting model for the entire dataset.

\noindent\textbf{Proposed methods contrasts.} Our research focuses on tree-based models for vector forecasts, despite their conventional use for scalar output. We propose a model collection setup to adapt these models for vector tasks. Tree-based models offer robust feature engineering opportunities, effectively utilizing multivariate data. A prior study showed tree-based models outperform statistical methods in forecast accuracy \citep{SVOBODA2021119430}. Additionally, we conducted deep learning experiments to validate the advantages of tree-based models, allowing for a comparison of different methods.

%

\subsubsection{Results analysis}
\noindent\textbf{Research context.} Related research typically focuses on overall forecasting accuracy, with models trained only once. Although this approach offers insight into general model performance, it often neglects time-related patterns in the data. Additionally, results analysis usually examines only classical properties, such as statistical aspects of error distribution by day or month, assuming a static forecasting model.


\noindent\textbf{Proposed methods contrasts.} While we cover standard analyses like overall forecasting accuracy and classical error distribution, our research emphasizes online learning. We investigate how forecast accuracy changes over time and how the number of detected change points affects predictions. Specifically, we focus on forecast details near these change points to understand how well our models adapt to evolving natural gas consumption or electricity load patterns.

\subsubsection{Forecast accuracy}
\noindent\textbf{Research context.} Given the novel dataset and the focus of related energy forecasting research on offline learning and single-value-ahead forecasts, direct numerical comparisons with existing results could be misleading. To address this, we have implemented specific measures to ensure the robustness of our results. Based on our previous research \citep{SVOBODA2021119430}, the Symmetric Mean Absolute Percentage Error (SMAPE) for 24-hour ahead forecasts for this dataset generally ranges of~$7\%$ to~$15\%$, depending on the method and its complexity.


\noindent\textbf{Proposed methods contrasts.} To address the unique challenges of our dataset and the lack of prior research on continual learning, we established several baseline models. These baselines served as reliable benchmarks for comparison against more advanced approaches. By integrating change point detection into our methods, we reduced the SMAPE by approximately 5 to 6.5 \% compared to these baselines. Importantly, this enhancement was achieved with minimal additional computational overhead, highlighting the efficiency of our approach in improving forecast accuracy.

 
\section{Experiments} \label{sec:experiment}
The experiments that were conducted aim to verify the accuracy and efficiency of the proposed method on real-world data and evaluate the defined model collection selection schemas as beneficial over naive baseline models. We intend to answer the following research questions:
\begin{itemize}
    \item \textbf{RQ1:} How does the performance of a single-model continuous learning approach (SMCA) compare to a multimodel approach aggregated by year quarter (QDMDC) in the context of real-world data?
    \item \textbf{RQ2:} May we benefit from the model aggregation scheme based on change points detected by the PELT algorithm compared to the quarter aggregation?
    \item \textbf{RQ3:} Does the number of detected change points affect the accuracy of the forecast?
    \item \textbf{RQ4:} Is there an advantage in employing forecast ensembling (MCPDMC-WA) or model switching (MCPDMC-SW) in proximity to change points, with the main consideration being the Symmetric Mean Absolute Percentage Error (SMAPE) metric?
\end{itemize}

\subsection{Experiment setup}
\begin{figure}[ht!]
  \centering
    \includegraphics[width=\linewidth]{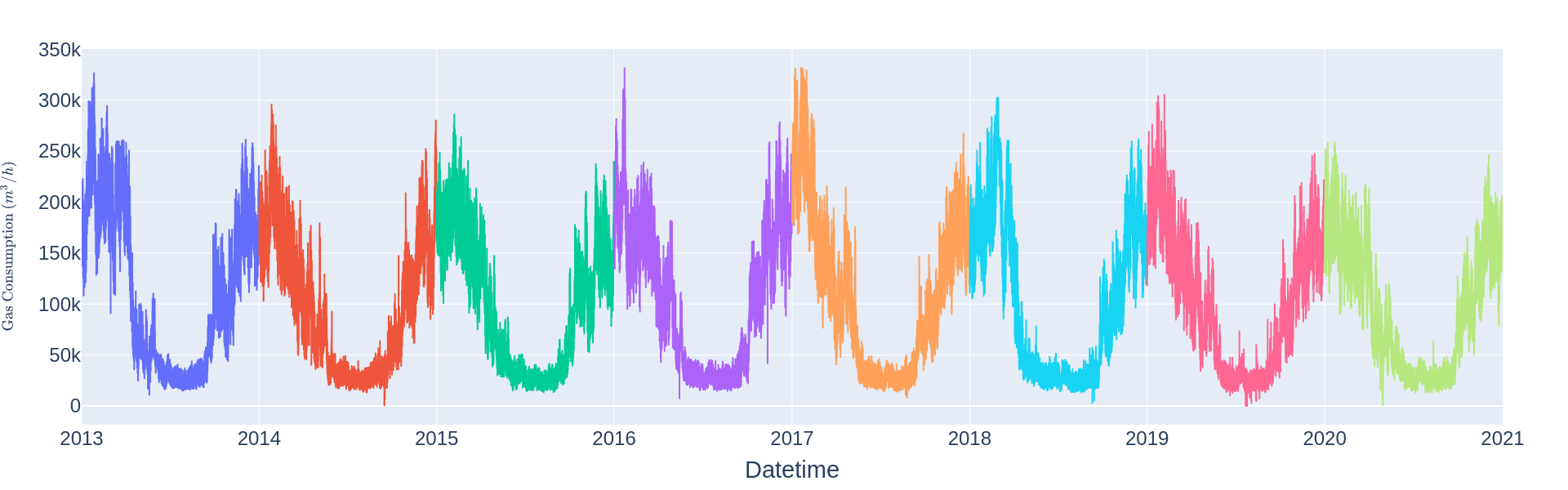}
  \caption{Natural gas consumption (endogenous variable) from January 1, 2013, to December 31, 2020 (years are divided by colors).}
  \label{fig:input_data_gas}
\end{figure}

\noindent\textbf{Dataset.} 
The data covers eight years, from January 1, 2013, to December 31, 2020, with data available at an hourly frequency~\citep{SVOBODA2021119430}\footnote{The dataset is available online: \url{https://ai.vsb.cz/natural-gas-forecasting}}. It consists of 70,104 data points, compiled from three main components. The first component comprises consumption data, focusing on Prague, the capital of the Czech Republic. The Prague distribution network served 422,926 customers in 2018, with a total consumption of 3.82 billion m3. The second component incorporates weather variables obtained from the Prague LKPR airport weather station. These variables are derived from periodic METAR reports issued by airports and are archived for long-term preservation. The third component that represents economic characteristics is natural gas price data. We have obtained price data from the Czech Energy Regulation Office and included them in the dataset. The natural gas consumption time series is shown in Fig.~\ref{fig:input_data_gas}, as well as the temperature, which is the most important exogenous variable, since gas consumption is highly dependent on the outside temperature because of the household heating.

\begin{figure}[ht!]
  \centering
    \includegraphics[width=\linewidth]{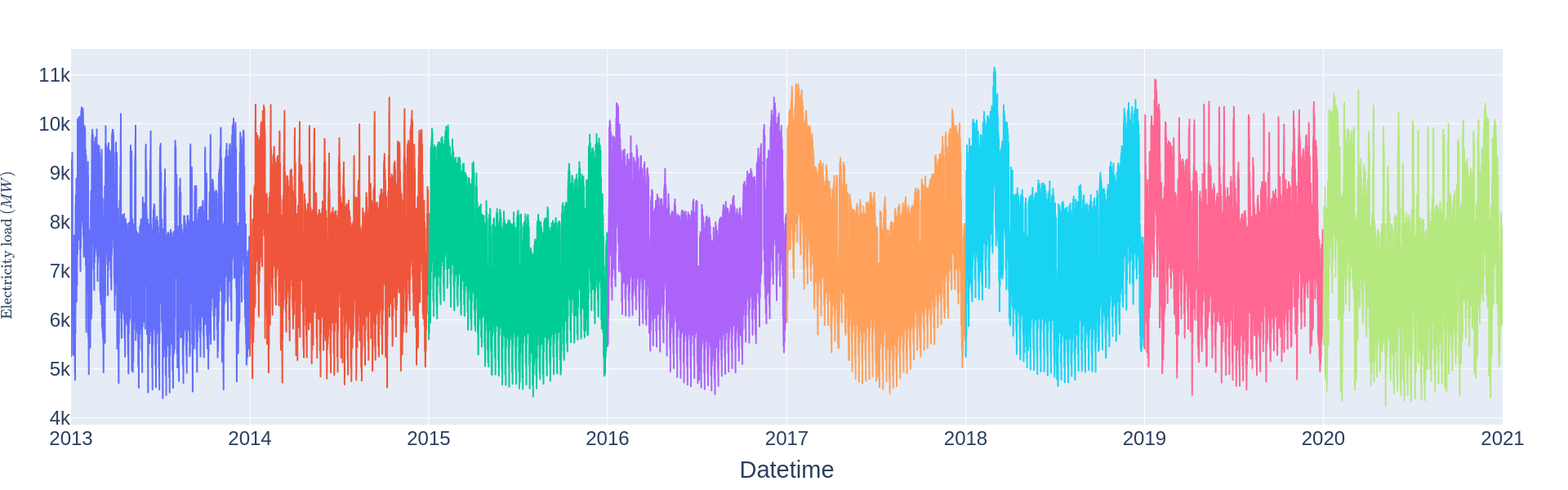}
  \caption{Electricity load (endogenous variable) from January 1, 2013, to December 31, 2020 (years are divided by colors).}
  \label{fig:input_data_el}
\end{figure}



Natural gas consumption over multiple years exhibits a clear cyclical phenomenon, as shown in Fig.~\ref{fig:input_data_gas}. Cycles reflect seasonal variations in outside temperature, affecting heating and cooling demand as the consumption is high during the cold and low during the warm months. Consumption also shows short-term trends corresponding to transition periods between the heating and summer seasons, when the temperature fluctuates rapidly.

To investigate the ability of the proposed method to generalize, we have conducted experiments also for the electricity load data. The electricity load forecasting experiments utilized a dataset comprising raw load data spanning the same time frame as the natural gas dataset, weather data and calendar information were sourced from the natural gas dataset, as they were measured at the same location. Fig. \ref{fig:input_data_el} depicts the electricity load time series. The detailed overview of the features of the dataset can be seen in \ref{sec:appxA} Table \ref{tab:features}.

\begin{figure}[!htb]
     \centering
     \begin{subfigure}{\linewidth}
         \centering
         \includegraphics[width=\linewidth]{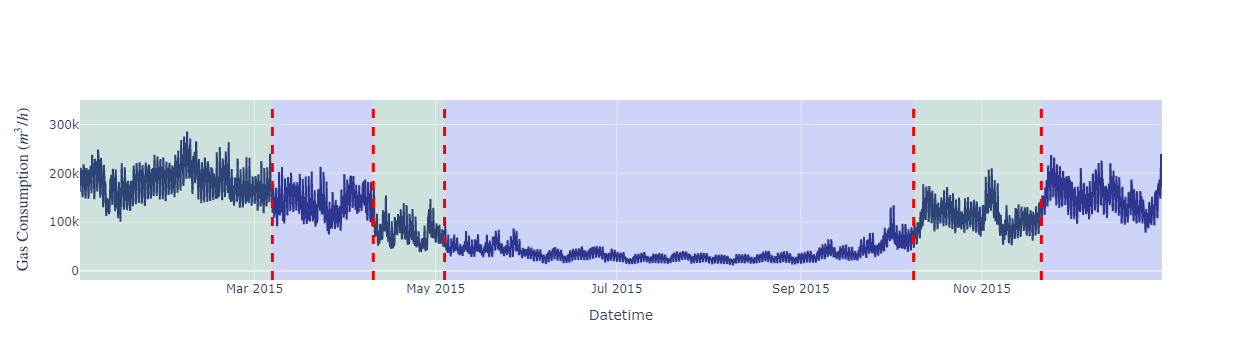}
         \caption{Change points detected in 2015 data.}
         \label{fig:cp_2015}
     \end{subfigure}
     \begin{subfigure}{\linewidth}
         \centering
         \includegraphics[width=\linewidth]{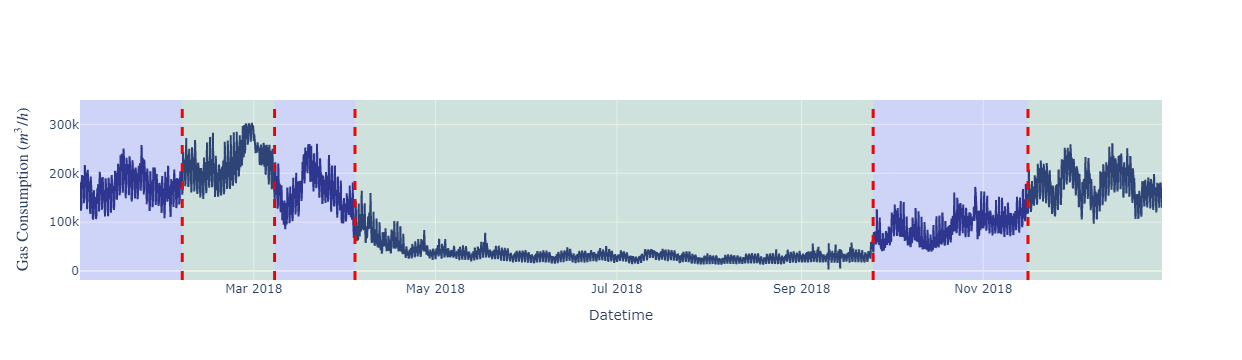}
         \caption{Change points detected in 2018 data.}
         \label{fig:cp_2018}
     \end{subfigure}
        \caption{Detected change points by PELT algorithm with \textit{Low} settings in selected years of the natural gas consumption data.}
        \label{fig:cp_examples}
\end{figure}

\begin{figure}[!htb]
     \centering
     \begin{subfigure}{\linewidth}
         \centering
         \includegraphics[width=\linewidth]{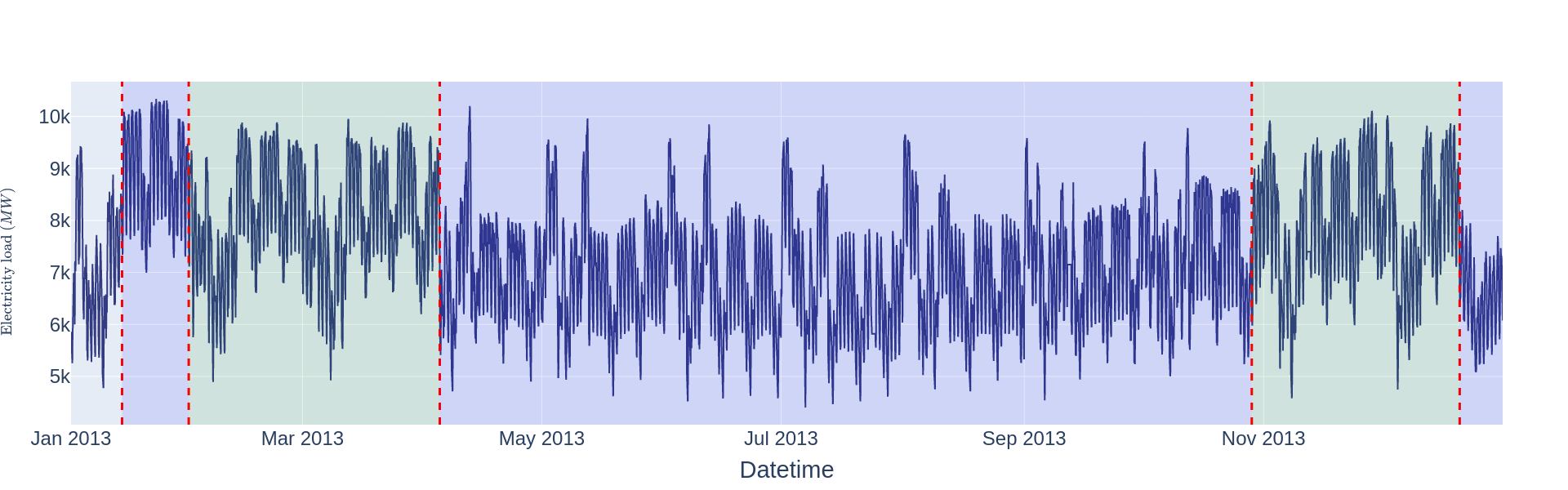}
         \caption{Change points detected in 2013 data.}
         \label{fig:cp_2013_el}
     \end{subfigure}
     \begin{subfigure}{\linewidth}
         \centering
         \includegraphics[width=\linewidth]{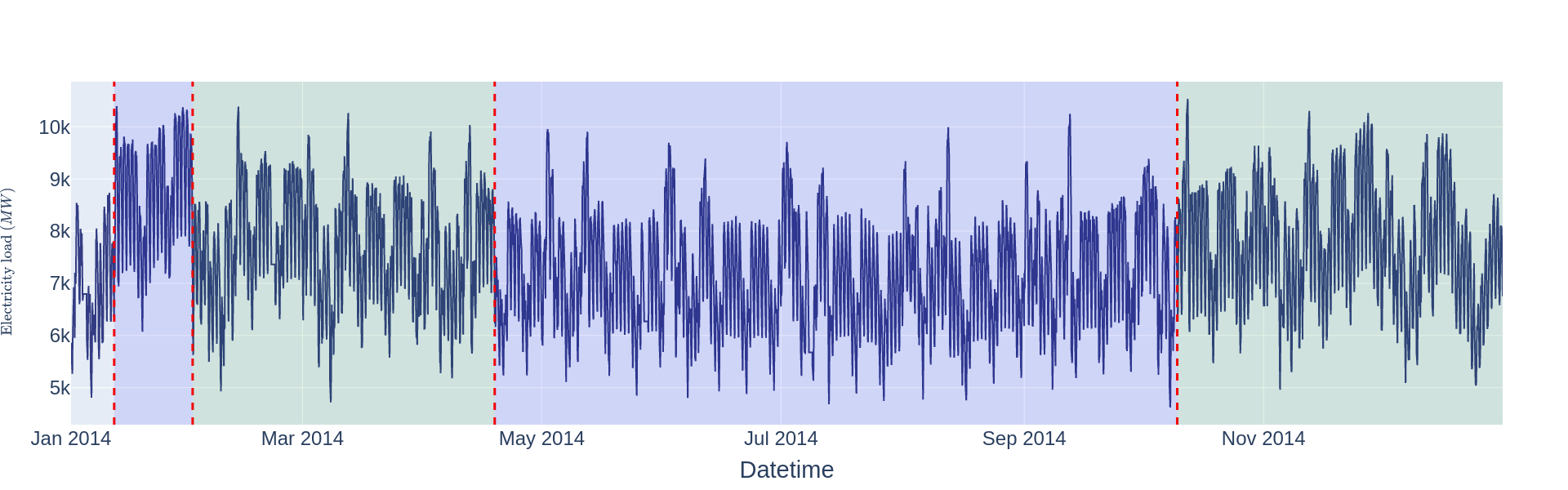}
         \caption{Change points detected in 2014 data.}
         \label{fig:cp_2014_el}
     \end{subfigure}
        \caption{Detected change points by PELT algorithm with \textit{Low} settings in selected years of the electricity load data.}
        \label{fig:cp_examples_el}
\end{figure}


Fig.~\ref{fig:cp_examples} illustrates the change points detected by the PELT algorithm (\textit{Low} settings) for selected years of natural gas consumption data. These change points divide the data into segments with varying levels or trends. During summer and fall, there is better segment overlap as consumption increases, whereas spring shows frequent level and trend changes, leading to lower overlap. The same analysis applied to electricity load data (see Fig.~\ref{fig:cp_examples_el}) reveals similar properties, but with evolving daily/weekly patterns, lower volatility, and greater stability.


\noindent\textbf{Dataset preprocessing.} The original data is prepared for machine learning by forming pairs of input and output vectors. Each input-output tuple corresponds to a forecast origin (e.g., midnight) in the time series. The input vector includes lagged features—historical values up to the forecast origin—such as Consumption (endogenous variable), and exogenous variables like Temperature, Wind Speed, or Humidity. These lagged variables are derived from a fixed window of 72 hours. Additionally, the input vector incorporates a Temperature Forecast for the next 24 hours\footnote{Forecasted by YR.no (\url{https://www.yr.no})}, along with standard date-time variables such as Day of the week, Month, and bank holiday indicators. Forecasts are made over a 24-hour horizon starting from midnight, with the output vector containing Consumption values for each hour of the subsequent day.

\begin{figure*}[ht!]
  \centering
    \includegraphics[width=1.0\textwidth]{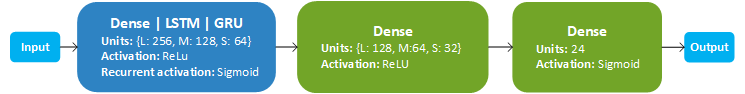}
  \caption{Deep learning model architecture used in the experiments. The scheme
shows the general architecture, the first layer can be either Dense, LSTM or GRU layer, it depends on the selected variant of the model. Three sizes of the deep learning model were tested, size of the network affects the number of neurons in the first two layers.}
  \label{fig:ann_architecture}
\end{figure*}

\noindent\textbf{Algorithms used in the forecasting pipeline.} The methodology outlined in Section~\ref{sec:methodology} initially employs a PELT algorithm for change point detection, chosen due to its established effectiveness in this domain (see Section~\ref{sec:rel_works} for details). For forecasting, the Hoeffding Tree is utilized because of its native support for incremental learning and strong performance of tree-based algorithms in offline scenarios~\citep{SVOBODA2021119430}. The Hoeffding Tree Regressor with $7$ instances (one week of the input data), a leaf should observe between split attempts (the so-called grace period), decay of the model selector of $0.2$ and threshold ($\tau$) below which a split is forced to break ties of~$0.5$ is used as a forecasting model. Hyper-parameters were tuned using the first year of data. A multimodel direct forecasting approach is adopted for the 24-hour forecasting task, where each output value is predicted by a corresponding model among a sequence of 24 models. In addition, experiments with deep learning models were conducted to compare performance with Hoeffding Tree Regressors. Three variants of the deep learning model were tested, each with variations in layer types (fully connected LSTM or GRU) and sizes (large, medium, small), detailed in Fig.\ref{fig:ann_architecture}. For change point detection, the PELT algorithm is configured with an $L2$ segment model, a minimum segment length of 168 hours, and a subsampling rate of 24 hours. Complete hyper-parameter settings for the forecasting algorithms used are summarized in Table\ref{tab:hyperparameters}.

\begin{table}[!htp]
\caption{Hyper-parameter settings for the algorithms used in the forecasting pipeline.}
\begin{center}
\begin{tabular}{|l|r|}
\hline
\textbf{Algorithm/Parameter} & \textbf{Parameter Value}  \\
\hline
General Parameters & \\
\textit{Past lags of input data} & 72 hours \\
\textit{Temperature forecast steps} & 24 hours \\
\textit{Window width for SWITCH and WAVG procedure} & 14 days \\

\hline
Hoeffding Tree Regressor & \\
\textit{Num. of trees in collection} & 24 \\
\textit{Grace period} & 7 \\
\textit{Decay} & 0.2 \\
\textit{Threshold ($\tau$)} & 0.5 \\
\hline
PELT & \\
\textit{Segment model} & L2 \\
\textit{Minimum segment length} & 168 hours \\
\textit{Subsample rate} & 24 hours \\
\textit{Penalization (Natural gas - Low, Medium, High)} & $\left(732, 244, 122\right) \times 10^9$ \\
\textit{Penalization (Electricity - Low, Medium, High)} & $\left(100, 150, 250\right) \times 10^6$ \\
\hline
Deep Learning Model & \\
\textit{Num. of hidden layers} & 2 \\
\textit{Activation func. hidden layers} & ReLU \\
\textit{Num. of neurons in 1st layer - Small, Medium, Large} & (256, 128, 64) \\
\textit{Num. of neurons in 2nd layer - Small, Medium, Large} & (128, 64, 32) \\
\hline
\end{tabular}
\end{center}
\label{tab:hyperparameters}
\end{table}

The computational complexity of the suggested methodology depends mainly on the implementation of the Hoeffding tree regressor. In general, the Hoeffding Tree regressor is generally considered to be linear with respect to the number of training instances (n) and logarithmic concerning the number of features (d) \citep{domingos2000}. During the training phase, the algorithm incrementally builds a decision tree as new instances arrive. At each step, it evaluates the splitting criteria for each candidate attribute to determine the best split. This process involves computing information gain or other statistical measures to assess the quality of the splits. The preprocessing steps use the linear transformation of the attributes and do not add complexity. Forecasting using a deep learning approach is the most computationally expensive, especially for LSTM-based models. The computational complexity of Long Short-Term Memory (LSTM) networks depends on several factors, including the number of input features (n), the number of LSTM units or cells (m), the length of the input sequence (t), and the number of output units (k).
The computational complexity of a single LSTM cell operation (including all gates and computations of cell state) is typically $O(m^2 + m*n)$, where m is the number of LSTM units and n is the number of input features. If the input sequence has length t and there are k output units, the total complexity of a single forward pass is approximately $O(tm^2 + tmn + tm*k)$ \citep{Hochreiter1997}. For deeper LSTM networks with multiple layers, the computational complexity increases with the number of layers. If the network has L layers, the overall complexity becomes $O(Ltm^2 + Ltmn + Ltmk)$. 
The computational complexity of the PELT (Pruned Exact Linear Time) algorithm depends on several factors, including the length of the time series data (n), the number of change points detected (m), and the desired level of precision. The PELT algorithm is known for its computational efficiency, especially compared to other change-point detection algorithms. It operates in linear time complexity, meaning that its computational cost increases linearly with the input data size ~\citep{Killick:2012}. Our proposed methods add only a small part of the computational complexity, resulting in a solid improvement in the precision of the prediction.

\noindent\textbf{Experimental protocol.} 
Because we employ data stream processing, thus only the current data instance is available to the model at each training step. Models are trained incrementally daily. The interleaved-test-then-train evaluation~\citep{KRAWCZYK2017132} was used in all experiments. It works with a data stream of instances, where each instance is first used for the inference (testing phase) and then for the model training, allowing for the gradual refinement of model accuracy. Following this sequential evaluation process, the model consistently faces unseen samples. This approach eliminates the need for a separate holdout set for testing, ensuring optimal utilization of the available data~\citep{BifetGavaldaEtAl18}.
The following experiments were carried out:

\begin{itemize}
    \item \textbf{Single Model Collection Approach (Baseline).} Only one model collection is used for the whole data stream. The model collection consists of 24 forecasting models, each used for one hour of the day.
    \item \textbf{Quarter-Divided Model Collection.} Each quarter of the year has its model collection. The collection choice depends on the current quarter of the year.
    \item \textbf{Pure Change Point-Divided Model Collection.} The year is divided into multiple segments by the change points in gas consumption detected by the PELT algorithm. Each of the segments has its own collection of models. The change points are detected using only the first year of the data (2013), and the same change point locations are reused in the next years. When the change point occurs, the model collections are switched.
    \item \textbf{Mixed Change Point-Divided Model Collection.} The approach is the same as the pure one at its core; however, there is a 7-day period before and after each change point. Two strategies (WAVG and SWITCH) defined in Section~\ref{sec:methodology} for forecast ensembling in these periods are tested. During these 14-day long periods, both model collections are trained on the data. Outside of these periods, the process works the same way as in the Pure Change Point-Divided Model Collection case.
\end{itemize}

\begin{figure}[htp!]
  \centering
    \includegraphics[width=0.9\textwidth]{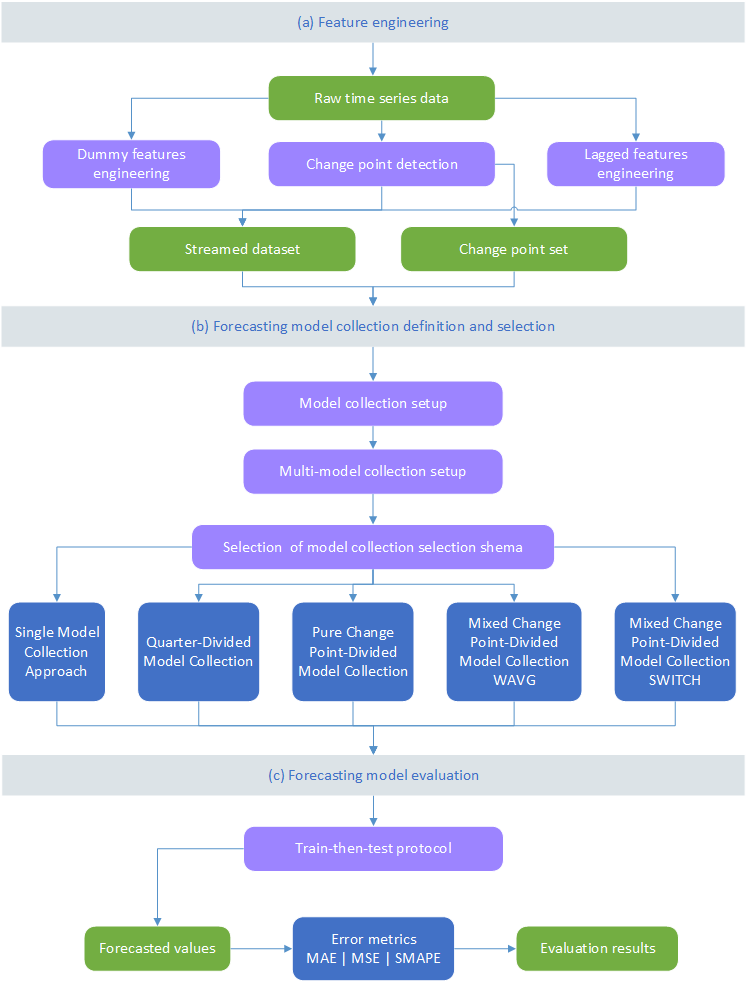}
  \caption{Designed experimental pipeline overview with high level of abstraction.}
  \label{fig:overall_pipeline}
\end{figure}

The Pure and Mixed Change Point-Divided Model Collection experiments were carried out with three different PELT algorithm settings, which detected 4 (\textit{Low} settings), 7 (\textit{Medium} settings), and 13 (\textit{High} settings) change points; to be able to assess the effect of increasing the number of detected change points (i.e. number of model collections) on the forecast error. The limitation of the PELT algorithm settings to only three options in our study was based on practical considerations. It is possible to introduce a more varied range of settings for the PELT algorithm, but in reality, the differences among various settings of the PELT algorithm might be relatively small, especially concerning the impact on performance, thus we adopted a scenario-based approach instead of exhaustively exploring a wide range of settings. This approach allowed us to effectively evaluate the algorithm's performance under different configurations without overly complicating the analysis, and our approach provided insights into its behavior and effectiveness using the three defined scenarios nonetheless. Fig. \ref{fig:overall_pipeline} shows the experimental setting we designed to validate the proposed methodology.

\noindent\textbf{Result evaluation.} 
Let $G_t$ be the ground truth of natural gas consumption or electricity load and the variable $F_t$ the forecast value, both at time step~$t$. We have used the standard set of commonly used error metrics, each summarizing different information, \citep{PETROPOULOS2022705}, \textit{MAE, MSE} and \textit{SMAPE} defined in expression~(\ref{eq:metrics_mae}),~(\ref{eq:metrics_mse}) and~(\ref{eq:metrics_smape}), respectively together with a Diebold-Mariano test \citep{Diebold2002}. 
\begin{equation} \label{eq:metrics_mae}
MAE = \frac{\sum_{t=1}^{n}|G_t - F_t|}{n},
\end{equation}
\begin{equation} \label{eq:metrics_mse}
MSE = \frac{\sum_{t=1}^{n}(G_t - F_t)^2}{n},
\end{equation}
\begin{equation} \label{eq:metrics_smape}
SMAPE = \frac{100}{n} \sum_{t=1}^{n} \frac{\mid G_t - F_t \mid}{\frac{1}{2} \left( \mid G_t \mid + \mid F_t \mid \right)}.
\end{equation}

The results are evaluated for the years 2014 to 2020. The first year (2013) was intentionally left out of the evaluation process because the data were used in the change point detection phase for the Pure and Mixed Change Point-Divided Model Collection experiments; thus, it could be taken as an information leak, and this phenomenon would distort the comparison with other approaches in this year.

\noindent\textbf{Implementation and reproducibility.} The experiments were carried out in Python (v. 3.11) using the Ruptures library (v. 1.1.7) for the PELT algorithm and the River library (v. 0.15.0) for the Hoeffding Tree Regressor. Tensorflow2 (v. 2.15.0) framework was used to train the deep learning models. Pandas (v. 1.5.3) and Numpy (v. 1.24.2) libraries were utilized for dataset manipulation and preprocessing. The codebase with the method and
experiment implementation is available in the git repository\footnote{\url{https://github.com/rasvob/Hoeffding-Trees-with-CPD-Multistep-Forecasing}}


\section{Results}
Two baseline and three models based on the proposed methodology were experimentally evaluated with three deep learning model architectures in three different settings to compare the Hoeffding Tree-based models to other approaches. Both baseline models omit the change point detection integration, as the first uses only one model collection for the data stream, and the second aggregates the model collections using year quarters. Models based on the proposed methodology were evaluated using three levels (low, medium, and high) of change point density. For each model, a set of selected error metrics were calculated.
The aggregated results are summarized in Table \ref{tab:el_general_res_selected} and detailed results are provided in Tables \ref{sec:appxB} \ref{tab:gas_general_res}, \ref{tab:gas_general_res_low}, \ref{tab:gas_general_res_med}, \ref{tab:gas_general_res_high} for the natural gas consumption data and in Table \ref{tab:el_general_res_selected} with detailed results provided in Tables \ref{sec:appxB} \ref{tab:el_general_res}, \ref{tab:el_general_res_low}, \ref{tab:el_general_res_med}, \ref{tab:el_general_res_high} for the electricity load data. Figures \ref{fig:sape_years} and \ref{fig:sape_years_el} provide a detailed view of the results that provides insight into the model accuracy differences is included in Figures \ref{fig:detailed_results}, \ref{fig:detailed_results_el} and Tables \ref{sec:appxB} \ref{tab:sape_by_month}, \ref{sec:appxB} \ref{tab:sape_by_month_el}.


\section{Lessons learned}
\label{Lessonslearned}
We have grouped conclusions according to the research questions formulated in Section~\ref{sec:experiment} to facilitate a systematic evaluation of the results obtained.


\begin{itemize}[\null]




\item \textit{\textbf{RQ1:} How does the performance of a single-model continuous learning approach (SMCA) compare to a multimodel approach aggregated by year quarter (QDMDC) in the context of real-world data?}

For the natural gas consumption data, the HT-SMCA approach exhibited a marginally lower SMAPE than the HT-QDMDC approach, a baseline that segments the data by quarters of the year (Table~\ref{tab:gas_general_res_selected}). Analysis of Fig.~\ref{fig:sape_years} shows that this discrepancy in SMAPE was mainly due to the lower SMAPE of HT-SMCA in 2014, after the first two years of training. With more training data, the models performed similarly with negligible differences. Furthermore, the HT-QDMDC approach yielded a higher SMAPE but slightly lower MSE than the HT-SMCA approach (Table~\ref{tab:gas_general_res_selected}). However, the performance difference was not significant. Consequently, the straightforward HT-SMCA approach proved preferable to the HT-QDMDC approach, which naively segmented the data by quarters of years without capitalizing on inherent cyclical patterns in natural gas consumption. 

For the electricity load data, the HT-SMCA also has lower error values than the HT-QDMDC approach (see Table~\ref{tab:el_general_res_selected}). Fig. \ref{fig:sape_years_el} shows the progress of SAPE over the years, and we can see that the HT-SMCA approach was always superior to HT-QDMDC.

Deep learning-based models performed worse than HT-SMCA or HT-QDMDC in both datasets, regardless of the SMCA or QDMDC. Moreover, the LSTM or GRU models had even higher forecasting errors than the fully connected model. Using just a single model for the whole dataset (SMCA) led to lower errors than the QDMDC. Thus, we may see that deep learning models do not benefit from model switching, as in the case of the Hoeffding Tree-based models.

\item \textit{\textbf{RQ2:} May we benefit from the model aggregation scheme based on change points detected by the PELT algorithm (PCPDMC) compared to the quarter aggregation?}

In the case of the natural gas consumption dataset, the change point divided model aggregation scheme (HT-PCPDMC) outperformed both baselines by all three metrics. The simplest variant (Pure) achieved the best performance in almost all PELT settings, except for the \textit{low} setting, where it was comparable to the weighted average forecast ensemble variant (HT-PCPDMC-WA). However, the error of HT-PCPDMC-WA increased with the number of change points. Tables~\ref{sec:appxB}~\ref{tab:gas_general_res_low},~\ref{tab:gas_general_res_med},~\ref{tab:gas_general_res_high} and Figures~\ref{fig:sape_low} and~\ref{fig:sape_high} show the data and the evolution of SMAPE over time in full detail and we can also see this phenomenon in Table~\ref{tab:gas_general_res_selected}. 

In electricity load dataset experiments, the HT-PCPDMC scheme also provided lower errors than HT-QDMDC regardless of the used number or change points. However, \textit{Low} and \textit{Medium} change point numbers provided better results compared to \textit{High} PELT setting, as we can see in the Table \ref{tab:el_general_res_selected} (or Tables~\ref{sec:appxB} ~\ref{tab:el_general_res_low},~\ref{tab:el_general_res_med},~\ref{tab:el_general_res_high} in more detailed view), the progress of SMAPE in different PELT settings can be seen in Fig. \ref{fig:sape_years_el}. We can see that electricity load forecasting becomes more complex over time than natural gas consumption, as there is an error level shift in all models used in 2019 and 2020. Furthermore, we can see that the proposed HT-PCPDMC performed the best even after the error level shift occurred, and SMAPE for 2020 was the lowest in all three PELT settings.

The PCPDMC scheme, compared to QDMDC, was beneficial for the deep learning models as well, as the SMAPE was lower. We may see that the lower number of change points provides better results as the SMAPE increases with the number of change points. However, we can see that deep learning models performed better if a single model was used for the entire data stream.

Fig. \ref{fig:dm_gas_heatmap} illustrates the p-values resulting from the Diebold-Mariano test conducted on the selected models for natural gas consumption. It is evident that, except for the HT-PCPDMC model utilizing the \textit{medium} PELT settings, each model employing an aggregation scheme based on change points detected by the PELT algorithm exhibits a p-value lower than 0.05, when compared to both the baseline HT-QDMDC and deep learning-based models.

Similarly, the p-values derived from the Diebold-Mariano test for the electricity load dataset are presented in Fig. \ref{fig:dm_el_heatmap}. Here, the comparison indicates that, across the baseline deep learning models and HT-QDMDC, the p-values for the comparison of the other models selected remain below the threshold of 0.05, except for the HT-PCPDMC model utilizing the \textit{high} PELT settings. Thus, irrespective of the dataset analyzed, the null hypothesis that there is no difference in the accuracy of the forecast between two competing forecasting models can be rejected.


\begin{figure*}[ht!]
  \centering
    \includegraphics[width=0.8\textwidth]{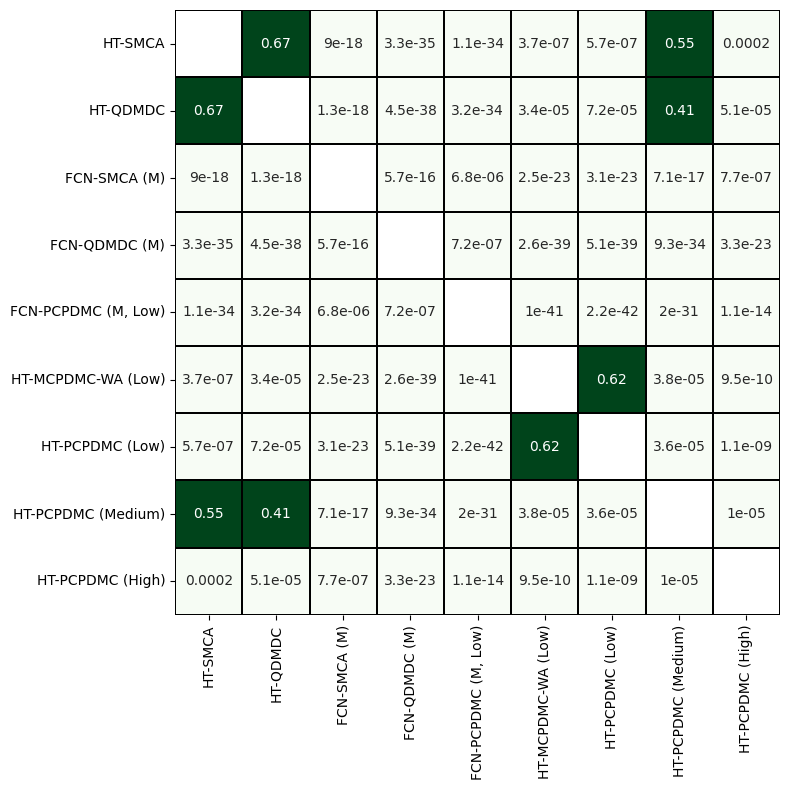}
  \caption{Matrix displaying $p-values$ of the Diebold-Mariano Test used for a pair-wise comparison of selection of the best performing models with the baseline approaches using both Hoeffding Trees and Deep learning models for the natural gas consumption dataset.}
  \label{fig:dm_gas_heatmap}
\end{figure*}

\begin{figure*}[ht!]
  \centering
    \includegraphics[width=0.8\textwidth]{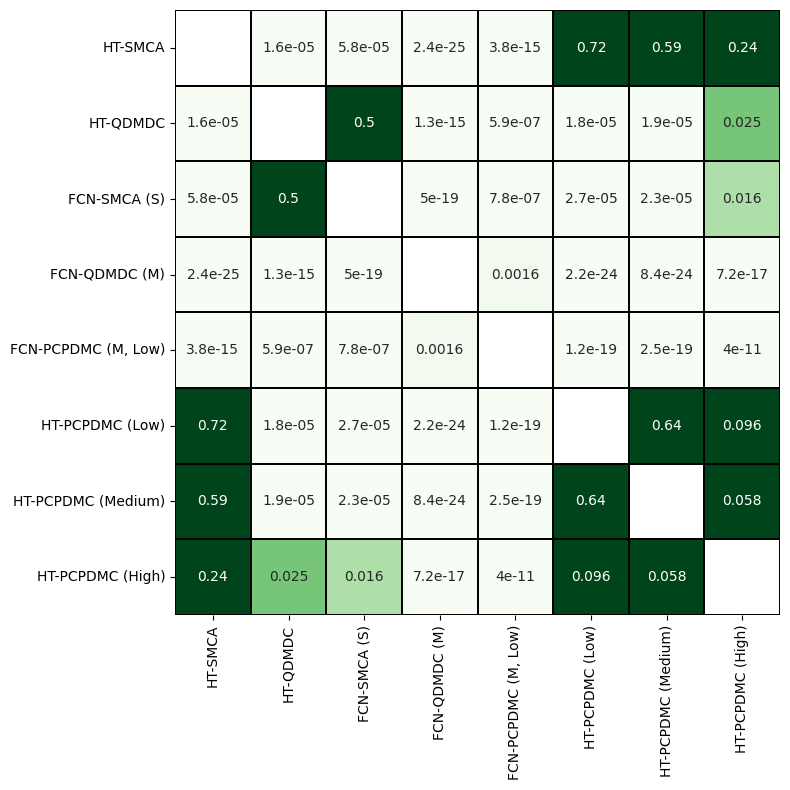}
  \caption{Matrix displaying $p-values$ of the Diebold-Mariano Test used for a pair-wise comparison of selection of the best performing models with the baseline approaches using both Hoeffding Trees and Deep learning models for the electricity load dataset.}
  \label{fig:dm_el_heatmap}
\end{figure*}

\item \textit{\textbf{RQ3:} Does the number of detected change points affect the accuracy of the forecast?}

The PELT settings had a significant effect on the accuracy of the forecast. 
The results of the natural gas consumption dataset indicated that fewer change points were preferable, as shown in Table~\ref{tab:el_general_res_selected}.
This is further corroborated by Tables~\ref{tab:gas_general_res_selected}~\ref{sec:appxB}~\ref{tab:gas_general_res_low},~\ref{tab:gas_general_res_med}, and~\ref{tab:gas_general_res_high}. 
More change points captured short-term changes in the data, often leading to overfitting of the model and increasing error. The \textit{Low} PELT setting yielded a lower forecast error for all collection selection schemes compared to the baselines, which was not the case for other setups. The \textit{High} PELT setting produced the highest forecast error for all approaches. This effect can be observed in Figures \ref{fig:pred_2020} to \ref{fig:sape_boxplot_2020}. The figures provide a comparison of the Pure Change Point-Divided Model Collection (HT-PCPDMC) with \textit{Low} PELT setting with both Mixed Change Point-Divided Model Collection approaches (HT-MCPDMC-WA and HT-MCPDMC-SW) but with \textit{High} PELT settings for the last forecasted year 2020. Figures \ref{fig:pred_2020} and \ref{fig:sape_2020} show that both HT-MCPDMC approaches with a higher number of change points have a bias towards forecasting values higher than the ground truth compared to the HT-PCPDMC approach together with a higher error variance, especially during the period from the end of March to the first half of April. To see the effect more clearly, Figure~\ref{fig:sape_boxplot_2020} and Table \ref{sec:appxB} \ref{tab:sape_by_month} show the aggregated errors by month; we may see that the median of the Symmetric Absolute Percentage Error in the majority of months is lower for the HT-PCPDMC approach compared to both HT-MCPDMC approaches and the biggest difference occurred during the mentioned period.
A detailed view of this period is shown in Fig. \ref{fig:pred_example_2020}. We may see that the underperformance of the HT-MCPDMC approaches is mainly caused by the model performance after April 7th as the models employing the HT-MCPDMC approach with a higher number of change points adapt to the slowly decreasing trend slower than the model based on the HT-PCPDMC approach with a lower number of change points.

The results for the electricity load data show that the lower number of change points was also preferable to a higher number. However, the results for \textit{Low} and \textit{Medium} The PELT settings show that for these settings, the results are similar, but \textit{High} number of change points used caused an increase in error. Figure~\ref{fig:pred_2020_el} shows that models using \textit{High} number of change points tend to overestimate the regular electricity load levels and if there is a peak in the load. We can also see in Figures~\ref{fig:sape_2020_el}~and~\ref{fig:sape_boxplot_2020_el}, and also Table \ref{sec:appxB} \ref{tab:sape_by_month_el}, that during the summer months the error levels are elevated of models utilizing \textit{High} number of change points compared to other models.

Deep learning models did not benefit from the higher number of change points either, and we can see in experiments using both experiments that a single deep learning model outperformed every other deep learning approach, the effect is more pronounced in the natural gas consumption datasets as in some cases the deep learning models using PCPDMC model selection approach even failed to converge (see Fig. \ref{fig:ngc_pred_fail_2020_v2}). Thus, the model switching approaches are suitable for the Hoeffding Tree-based models but not for the deep learning models, but we may observe in, e.g., Fig. \ref{fig:pred_example_2020_el} that even the single deep learning model (FCN-SMCA) without using any switching is not able to compare to the Hoeffding Tree-based models because the adaptability to changing levels of electricity consumption is lower.

\begin{figure*}[ht!]
  \centering
    \includegraphics[width=0.95\textwidth]{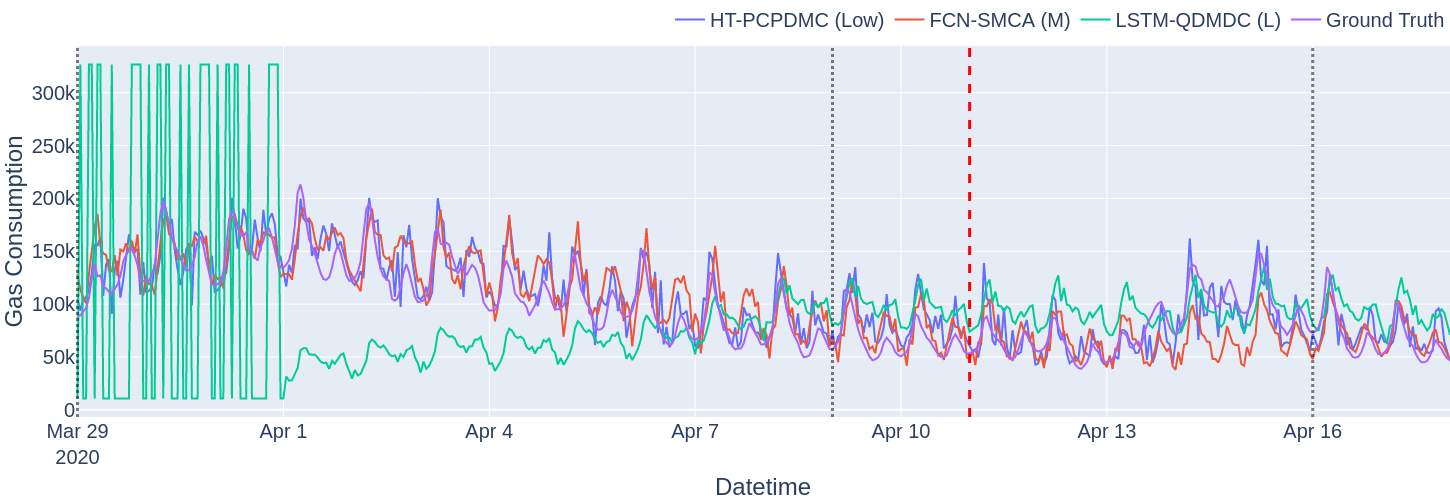}
  \caption{Example of an extreme behaviour of a deep learning model (LSTM-QDMDC (L)) which failed to coverge when there were different models used in different segments.}
  \label{fig:ngc_pred_fail_2020_v2}
\end{figure*}

\begin{figure}[!htp]
     \centering
     \begin{subfigure}{\linewidth}
         \centering
         \includegraphics[width=0.9\linewidth]{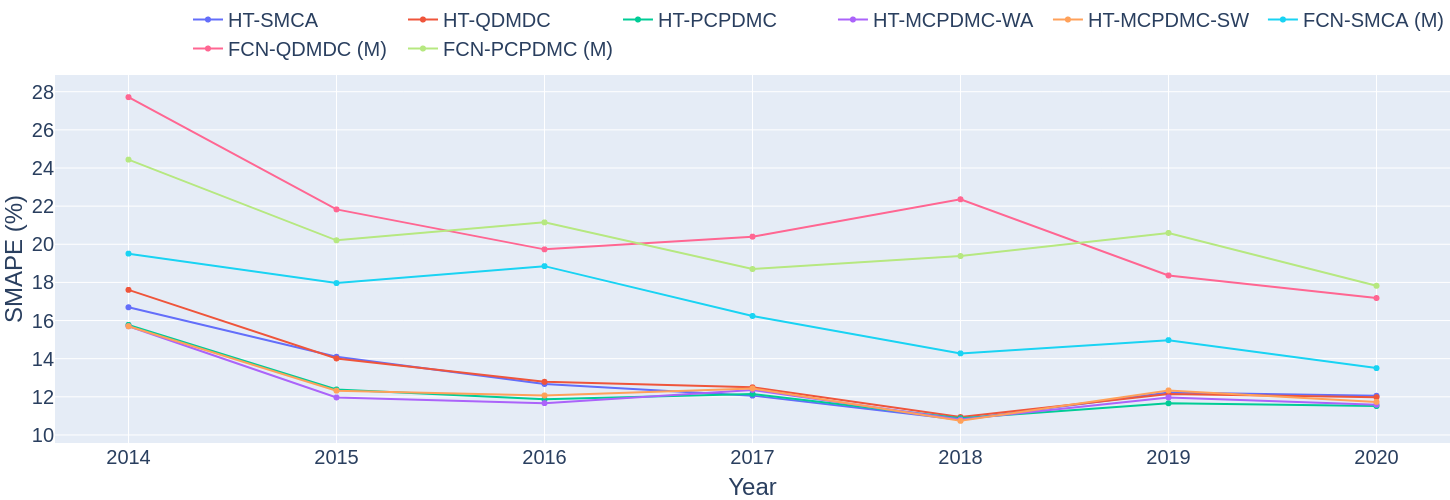}
         \caption{SMAPE yearly aggregated progress during test-then-train experiment protocol for PELT Settings = \textit{Low}.}
         \label{fig:sape_low}
     \end{subfigure}
     \begin{subfigure}{\linewidth}
         \centering
         \includegraphics[width=0.9\linewidth]{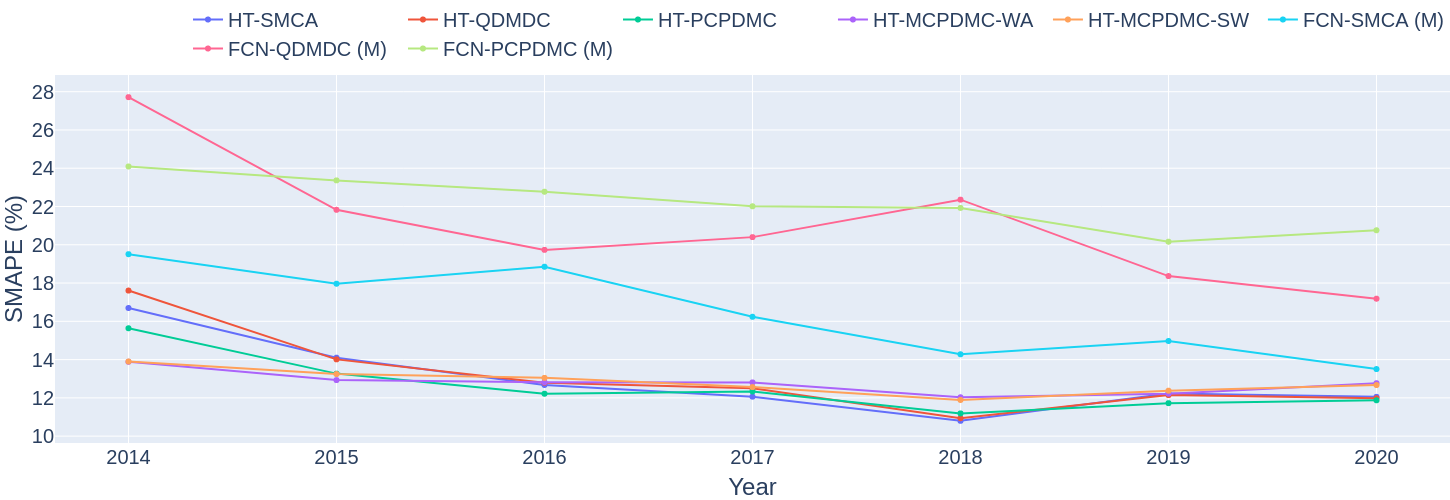}
         \caption{SMAPE yearly aggregated progress during test-then-train experiment protocol for PELT Settings = \textit{Medium}.}
         \label{fig:sape_med}
     \end{subfigure}
     \begin{subfigure}{\linewidth}
         \centering
         \includegraphics[width=0.9\linewidth]{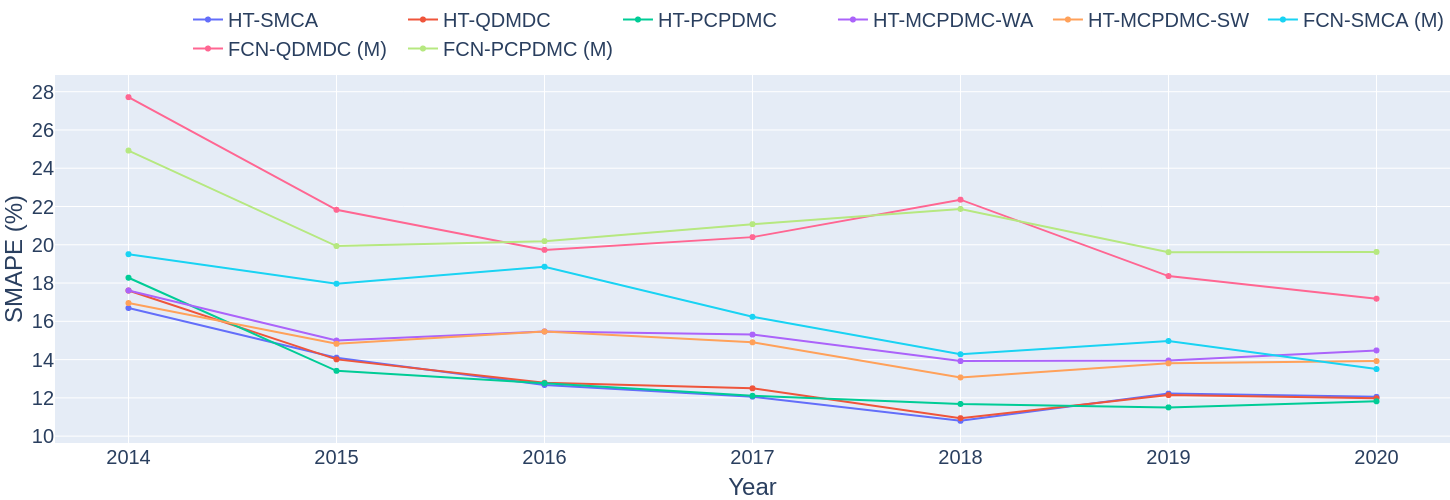}
         \caption{SMAPE yearly aggregated progress during test-then-train experiment protocol for PELT Settings = \textit{High}.}
         \label{fig:sape_high}
     \end{subfigure}
        \caption{Progress of yearly aggregated SMAPE for the conducted experiments using natural gas consumption dataset, divided by the change point detection algorithm setting, proposed models, and baseline comparison, is included for each setting. The SMAPE error measure was computed for the forecasted data from January 1, 2014, to December 31, 2020.}
        \label{fig:sape_years}
\end{figure}

\begin{figure}[!htp]
     \centering
     \begin{subfigure}{\linewidth}
         \centering
         \includegraphics[width=0.9\linewidth]{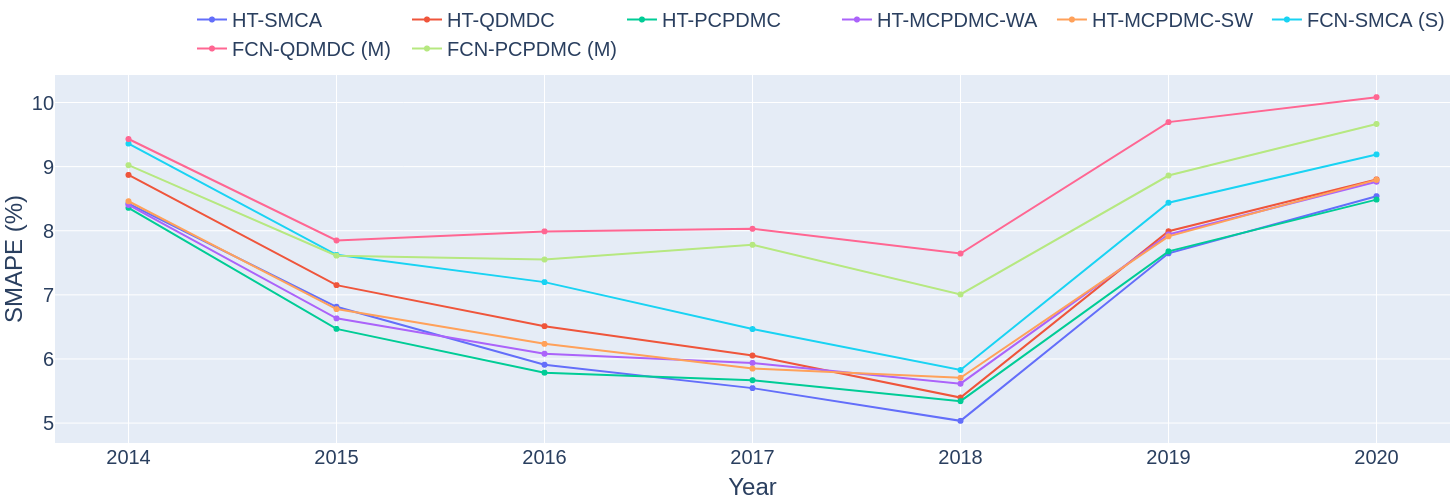}
         \caption{SMAPE yearly aggregated progress during test-then-train experiment protocol for PELT Settings = \textit{Low}.}
         \label{fig:sape_low_el}
     \end{subfigure}
     \begin{subfigure}{\linewidth}
         \centering
         \includegraphics[width=0.9\linewidth]{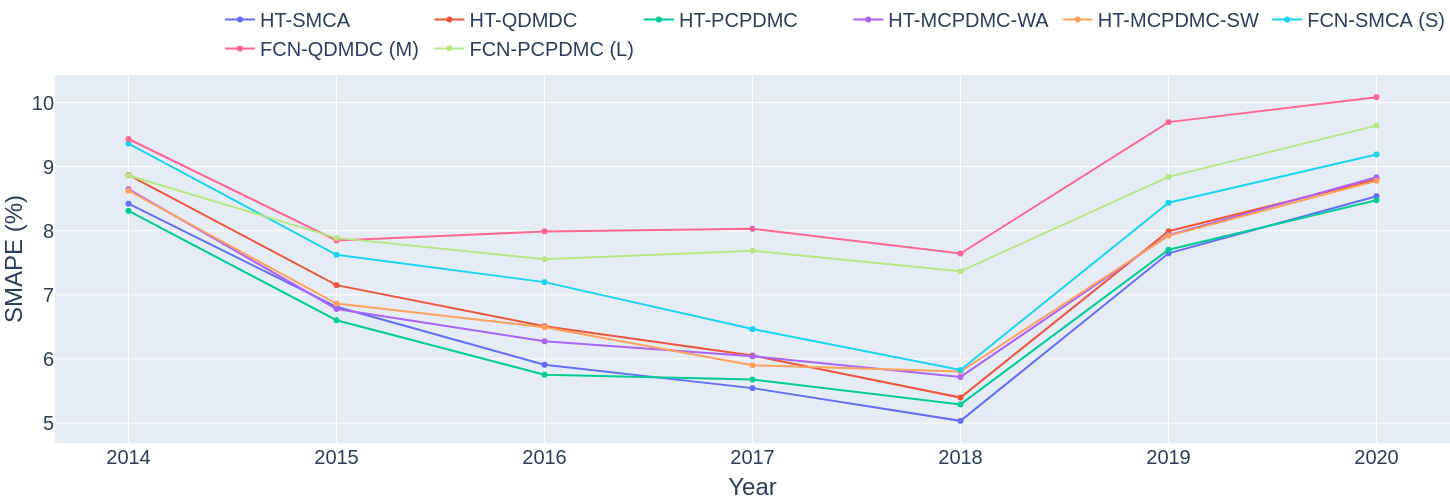}
         \caption{SMAPE yearly aggregated progress during test-then-train experiment protocol for PELT Settings = \textit{Medium}.}
         \label{fig:sape_med_el}
     \end{subfigure}
     \begin{subfigure}{\linewidth}
         \centering
         \includegraphics[width=0.9\linewidth]{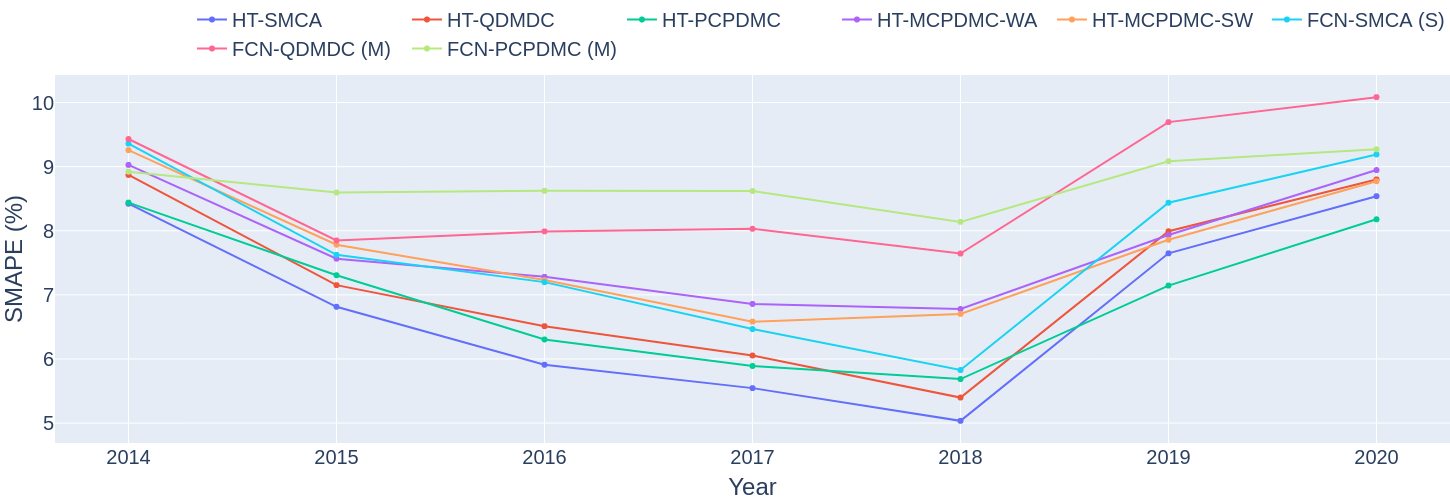}
         \caption{SMAPE yearly aggregated progress during test-then-train experiment protocol for PELT Settings = \textit{High}.}
         \label{fig:sape_high_el}
     \end{subfigure}
        \caption{Progress of yearly aggregated SMAPE for the conducted experiments using electricity load dataset, divided by the change point detection algorithm setting, proposed models, and baseline comparison, is included for each setting. The SMAPE error measure was computed for the forecasted data from January 1, 2014, to December 31, 2020.}
        \label{fig:sape_years_el}
\end{figure}

\begin{figure}[!htp]
    \centering
    \begin{adjustbox}{minipage=\textwidth,scale=0.75}
    \begin{subfigure}{\textwidth}
         \centering
            \includegraphics[width=1.0\textwidth]{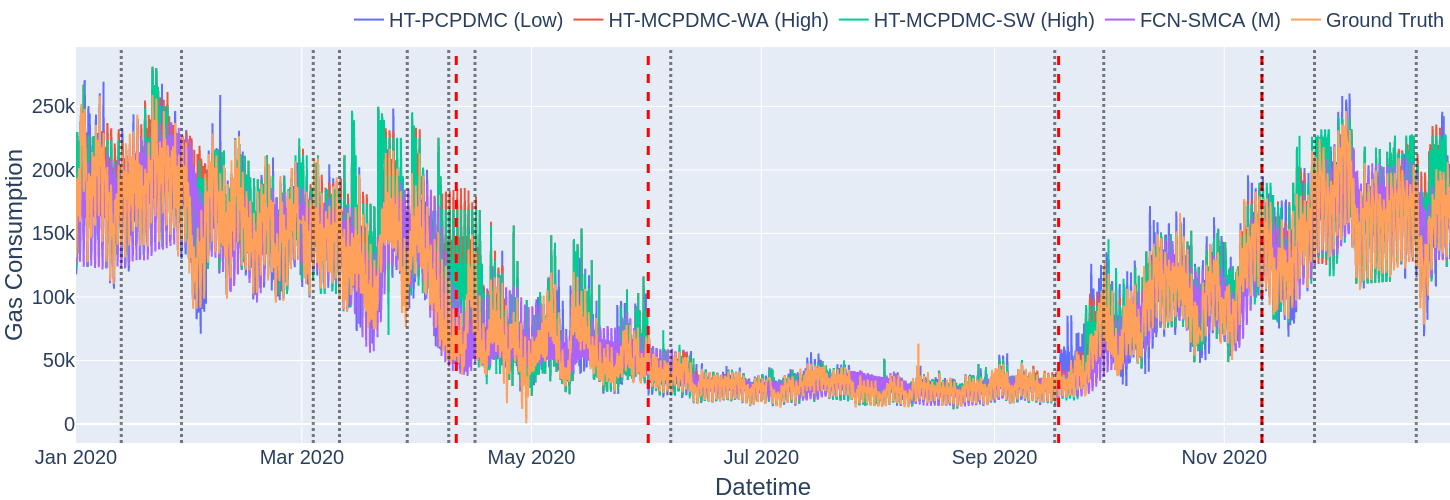}
          \caption{Ground truth natural gas consumption compared with forecasted values for 2020.}
          \label{fig:pred_2020}
     \end{subfigure}

     \begin{subfigure}{\textwidth}
         \centering
            \includegraphics[width=1.0\textwidth]{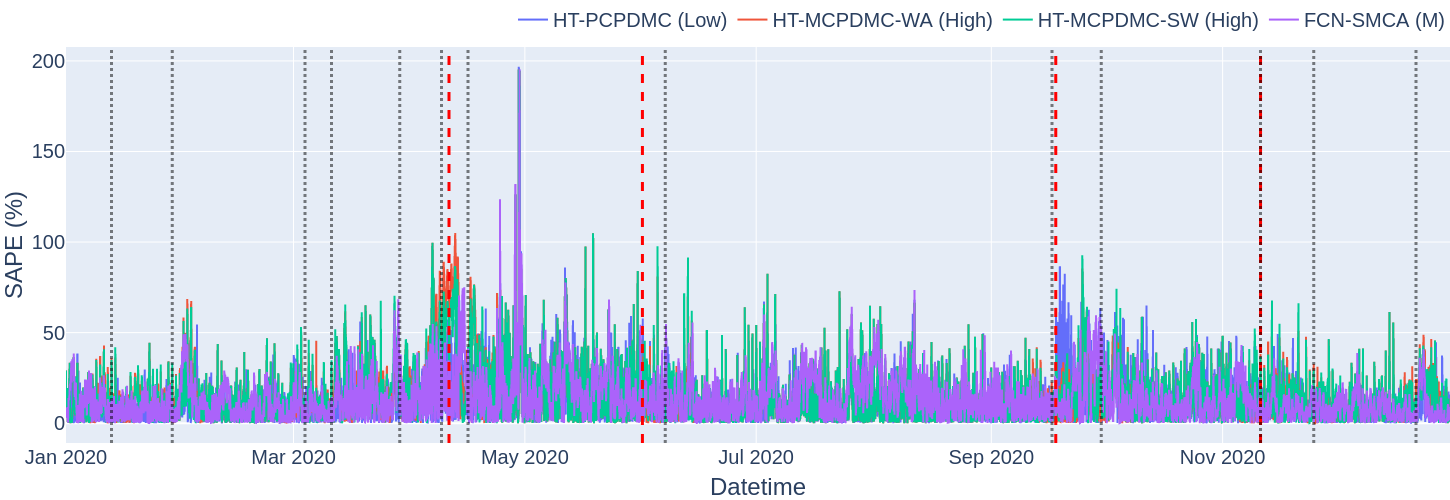}
          \caption{Symmetric Absolute Percentage Error (SAPE) plot of the selected forecasting approaches for 2020.}
          \label{fig:sape_2020}
     \end{subfigure}

     \begin{subfigure}{\textwidth}
            \centering
            \includegraphics[width=1.0\textwidth]{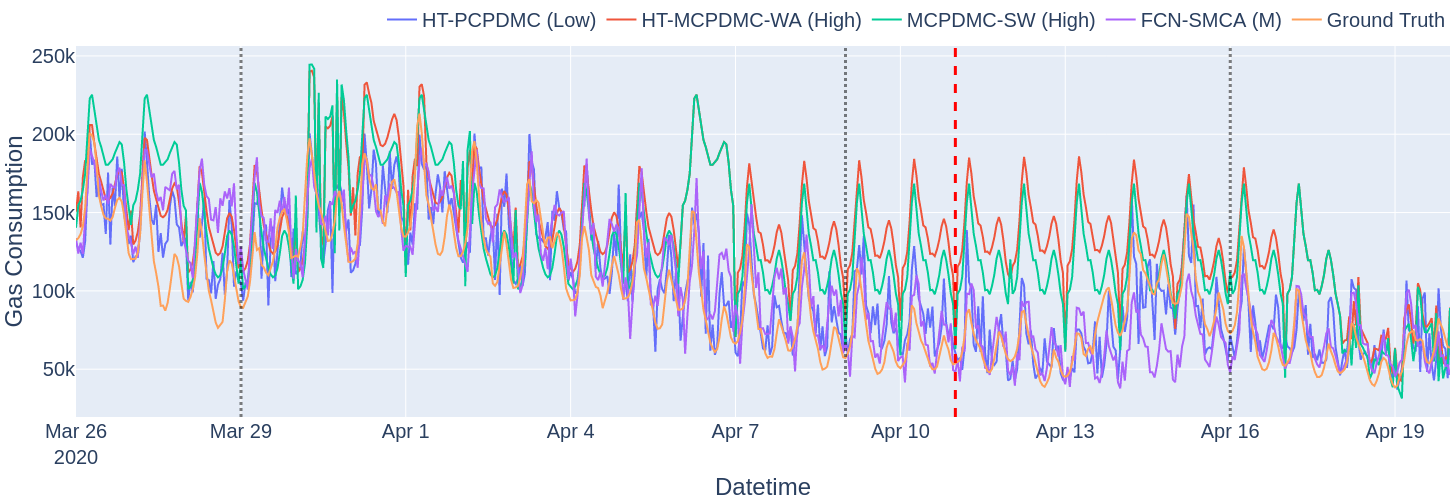}
          \caption{Ground truth natural gas consumption compared with forecasted values for a selected period in 2020, during which a multiple change points occur and impact in the forecasting accuracy.}
          \label{fig:pred_example_2020}
     \end{subfigure}

     \begin{subfigure}{\textwidth}
         \centering
            \includegraphics[width=1.0\textwidth]{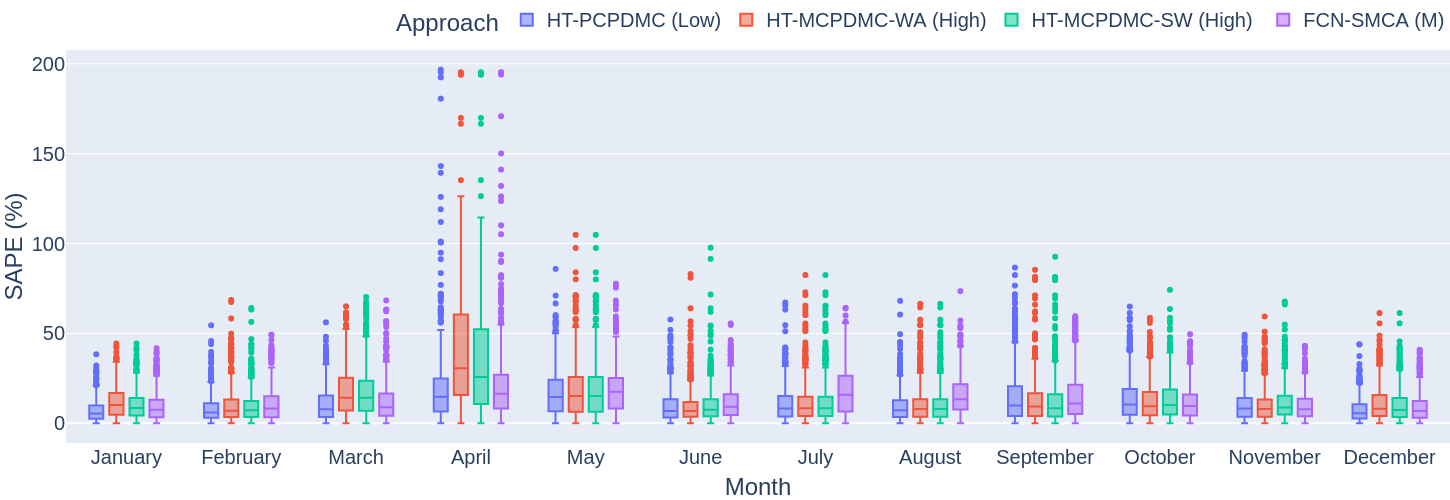}
          \caption{Symmetric Absolute Percentage Error (SAPE) of the selected forecasting approaches aggregated by months for 2020.}
          \label{fig:sape_boxplot_2020}
     \end{subfigure}
    \caption{Detailed results of HT-PCPDMC (\textit{Low} PELT settings), HT-MCPDMC-WA, HT-MCPDMC-SW (\textit{High} PELT settings for both) and FCN-SMCA (M) for the 2020 natural gas consumption data. The red dashed and black dotted vertical lines represent the change point locations detected with \textit{Low}, respectively \textit{High}, PELT algorithm settings.}
    \label{fig:detailed_results}
 \end{adjustbox}
\end{figure}

\begin{figure}[!htp]
    \centering
    \begin{adjustbox}{minipage=\textwidth,scale=0.75}
    \begin{subfigure}{\textwidth}
         \centering
            \includegraphics[width=1.0\textwidth]{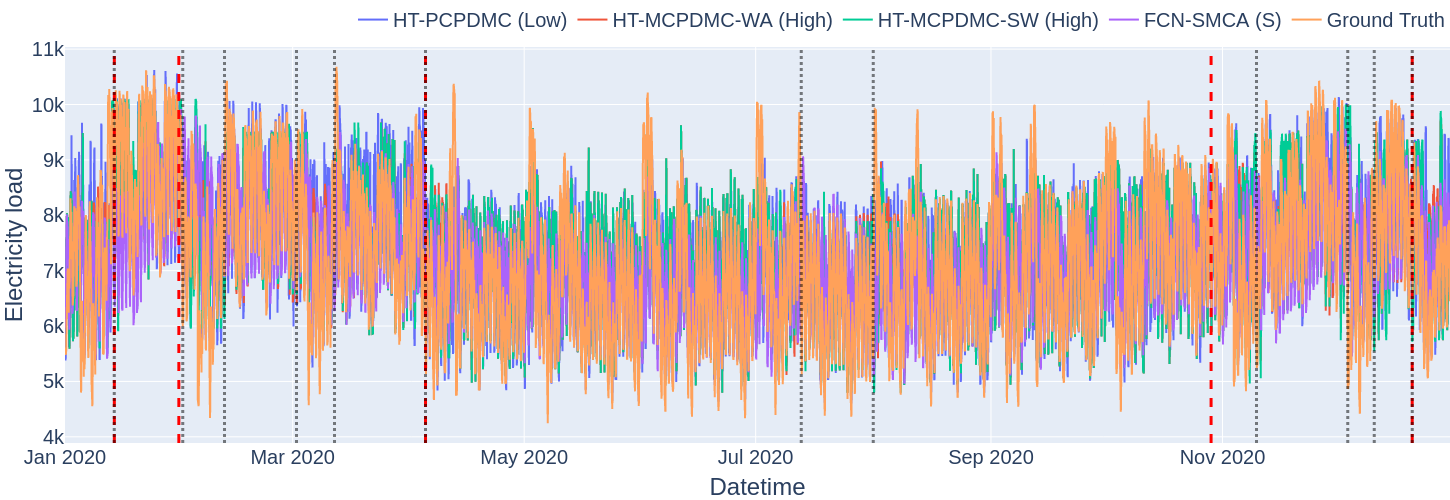}
          \caption{Ground truth electricity load compared with forecasted values using selected approaches for 2020.}
          \label{fig:pred_2020_el}
     \end{subfigure}

     \begin{subfigure}{\textwidth}
         \centering
            \includegraphics[width=1.0\textwidth]{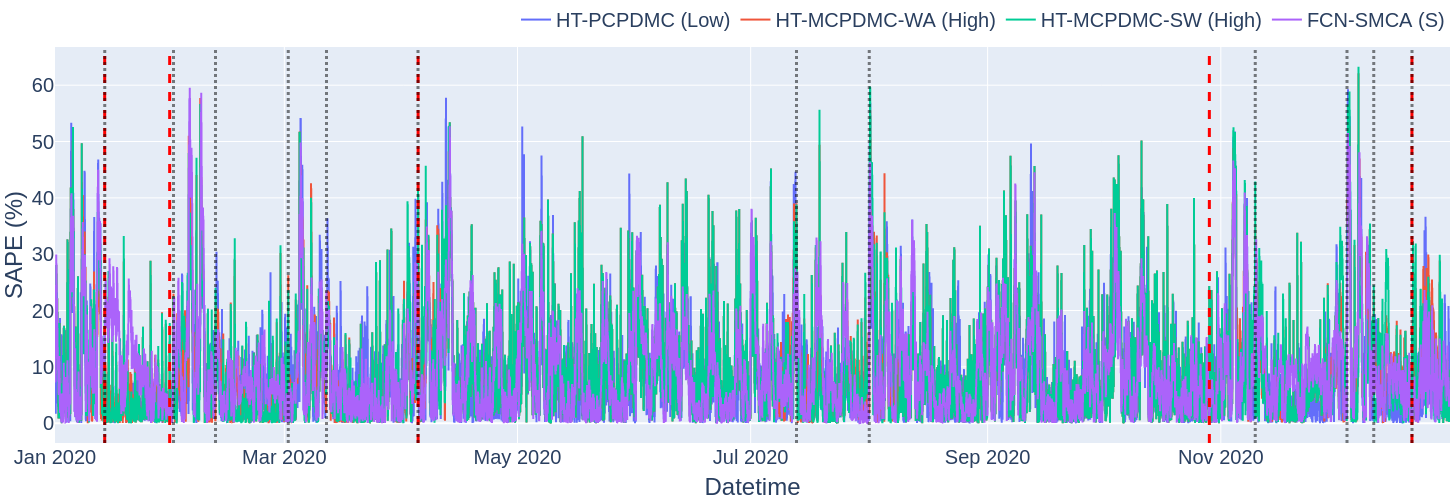}
          \caption{Symmetric Absolute Percentage Error (SAPE) plot of the selected forecasting approaches for 2020.}
          \label{fig:sape_2020_el}
     \end{subfigure}

     \begin{subfigure}{\textwidth}
            \centering
            \includegraphics[width=1.0\textwidth]{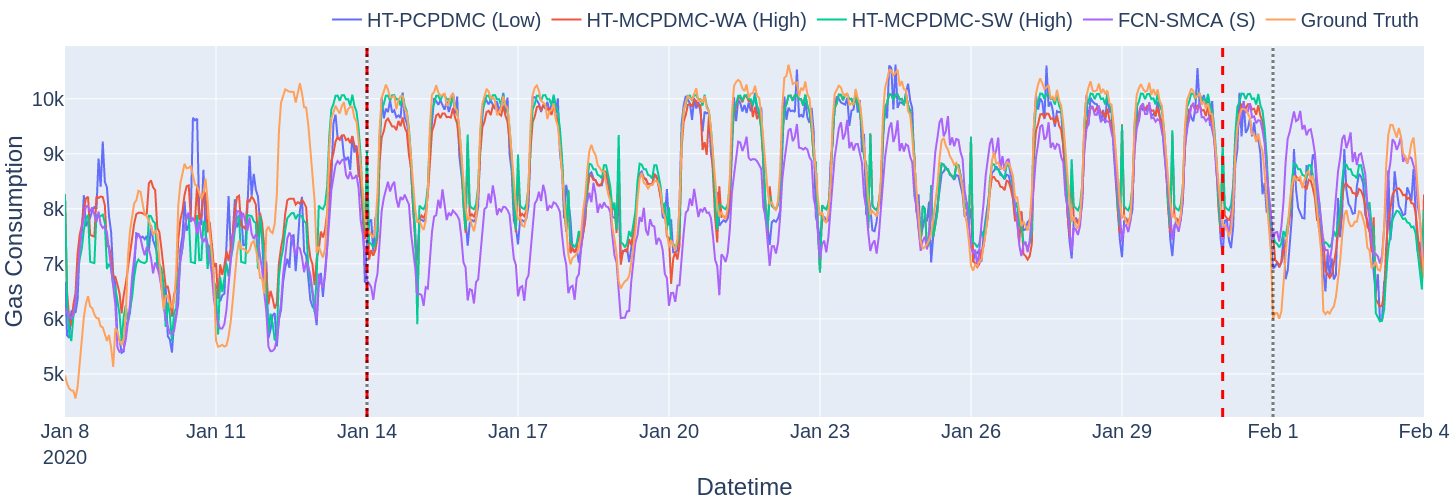}
          \caption{Ground truth electricity load compared with forecasted values using selected approaches for the selected part of the year 2020 in which a multiple change points occur and significant forecasting accuracy difference among the methods is present.}
          \label{fig:pred_example_2020_el}
     \end{subfigure}

     \begin{subfigure}{\textwidth}
         \centering
            \includegraphics[width=1.0\textwidth]{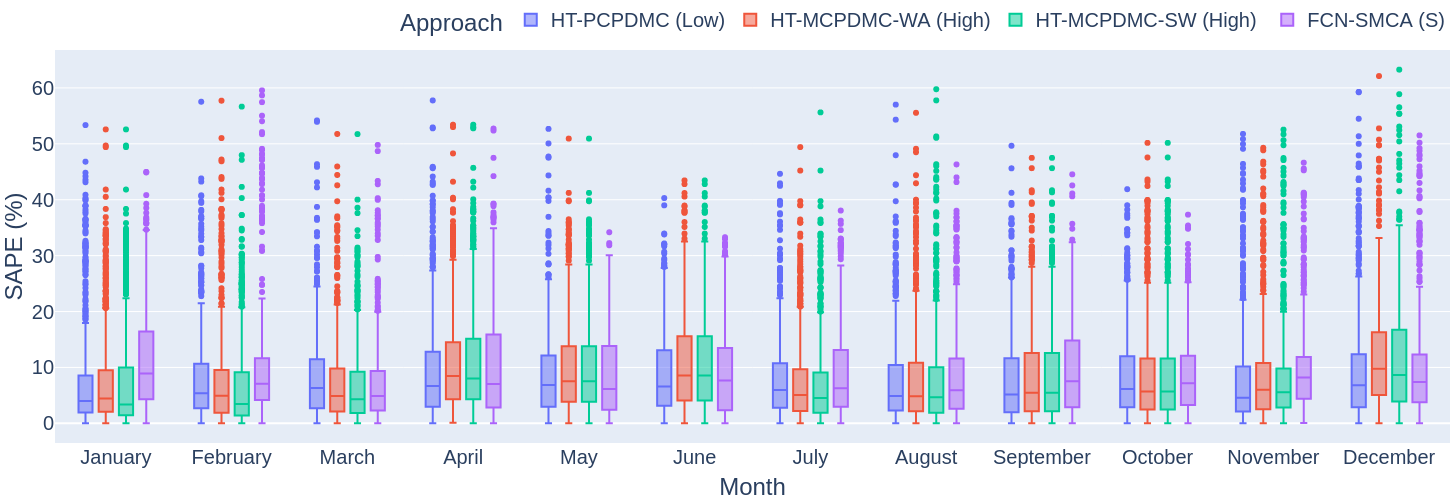}
          \caption{Symmetric Absolute Percentage Error (SAPE) of the selected forecasting approaches aggregated by months for 2020.}
          \label{fig:sape_boxplot_2020_el}
     \end{subfigure}
    \caption{Detailed look at results of HT-PCPDMC (\textit{Low} PELT settings), HT-MCPDMC-WA, HT-MCPDMC-SW (\textit{High} PELT settings for both) and FCN-SMCA (S) approaches for the 2020 electricity load data. The red dashed and black dotted vertical lines present in subfigures (a) to (c) represent the change point locations detected with \textit{Low}, respectively \textit{High}, PELT algorithm settings.}
    \label{fig:detailed_results_el}
 \end{adjustbox}
\end{figure}

\begin{table}[!htp]
\caption{Selected results of experiments conducted over natural gas consumption dataset using the baseline and proposed approaches. The MAE, MSE and SMAPE measures were computed for the forecasted data from January 1, 2014, to December 31, 2020.}
\begin{center}
\begin{adjustbox}{width=0.9\textwidth,center}
\begin{tabular}{|l|r|r|r|}
\hline
\textbf{Approach} & \textbf{MAE} & \textbf{MSE} & \textbf{SMAPE} \\
\hline
HT-SMCA              & 1.170e+04     & 3.143e+08    & \textbf{12.94}          \\
HT-QDMDC             & 1.179e+04    & 3.109e+08    & 13.14          \\
FCN-SMCA (M) & 1.548e+04 & 5.398e+08 &	16.49 \\
GRU-SMCA (L) & 1.798e+04 &	7.080e+08 &	19.03 \\
FCN-QDMDC (M) & 1.938e+04  & 	8.493e+08  & 	21.08 \\
GRU-QDMDC (M) & 2.306e+04  & 	1.109e+09  & 	27.50 \\
\hline
HT-PCPDMC (Low)            & 1.110e+04     & 2.783e+08    & 12.32          \\
HT-MCPDMC-WA (Low)       & 1.108e+04    & 2.770e+08     & \textbf{12.29}          \\
HT-MCPDMC-SW (Low)     & 1.124e+04    & 2.856e+08    & 12.48          \\
FCN-PCPDMC (M, Low) & 1.750e+04 &	6.457e+08 &	20.34  \\
LSTM-PCPDMC (L, Low) & 2.143e+04 &	9.764e+08 &	24.23  \\
\hline
HT-PCPDMC (Medium)        & 1.181e+04    & 3.211e+08    & \textbf{12.60}           \\
HT-MCPDMC-WA (Medium)     & 1.196e+04    & 3.233e+08    & 12.78          \\
HT-MCPDMC-SW (Medium)     & 1.189e+04    & 3.223e+08    & 12.81          \\
FCN-PCPDMC (L, Medium) & 2.010e+04 & 	8.929e+08	 & 22.08 \\
LSTM-PCPDMC (L, Medium) & 2.187e+04 & 	1.053e+09 & 	23.87  \\
\hline
HT-PCPDMC (High)            & 1.265e+04    & 3.925e+08    & \textbf{13.09}          \\
HT-MCPDMC-WA (High)       & 1.557e+04    & 6.368e+08    & 15.12          \\
HT-MCPDMC-SW (High)     & 1.483e+04    & 5.711e+08    & 14.72    \\
FCN-PCPDMC (M, High) & 1.975e+04 &	8.508e+08 &	21.05 \\
LSTM-PCPDMC (L, High) & 2.097e+04 &	9.366e+08 &	23.11  \\
\hline
\end{tabular}
\end{adjustbox}
\end{center}
\label{tab:gas_general_res_selected}
\end{table}

\begin{table}[!htp]
\caption{Selected results of experiments conducted over electricity load dataset using the baseline and proposed approaches. The MAE, MSE and SMAPE measures were computed for the forecasted data from January 1, 2014, to December 31, 2020.}
\begin{center}
\begin{adjustbox}{width=0.85\textwidth,center}
\begin{tabular}{|l|r|r|r|}
\hline
\textbf{Approach} & \textbf{MAE} & \textbf{MSE} & \textbf{SMAPE} \\
\hline
HT-SMCA              &      507.2 & 	5.053e+05 & 	\textbf{6.84}     \\
HT-QDMDC             &      537.6 & 	5.480e+05 & 	7.25     \\
FCN-SMCA (S) & 573.5 & 	5.571e+05 & 	7.73 \\
LSTM-SMCA (M) & 627.6	 & 6.698e+05 & 	8.49   \\
FCN-QDMDC (M) & 646.1	 & 6.733e+05 & 	8.68   \\
LSTM-QDMDC (L) & 665.4 & 	7.370e+05 & 	9.00  \\
\hline
HT-PCPDMC (Low)            & 506.5 & 	5.020e+05 & 	\textbf{6.83}         \\
HT-MCPDMC-WA (Low)       & 522.6 & 	5.344e+05 & 	7.05          \\
HT-MCPDMC-SW (Low)     & 526.6 & 	5.445e+05 & 	7.10          \\
FCN-PCPDMC (M, Low) & 611.6 & 	6.251e+05 & 	8.22  \\
LSTM-PCPDMC (L, Low) & 633.1 & 	6.768e+05 & 	8.56  \\
\hline
HT-PCPDMC (Medium)            & 506.2 &	5.001e+05 &	\textbf{6.83}         \\
HT-MCPDMC-WA (Medium)       & 532.1 &	5.467e+05 &	7.17          \\
HT-MCPDMC-SW (Medium)     & 533.3 &	5.555e+05 &	7.11         \\
FCN-PCPDMC (L, Medium) & 615.1 &	6.361e+05 &	8.26 \\
LSTM-PCPDMC (L, Medium) & 625.6 &	6.661e+05 &	8.46  \\
\hline
HT-PCPDMC (High)            & 517.2 & 	5.197e+05 & 	\textbf{6.99}         \\
HT-MCPDMC-WA (High)       & 576.0 & 	6.168e+05 & 	7.77          \\
HT-MCPDMC-SW (High)     & 573.1 & 	6.278e+05 & 	7.74         \\
FCN-PCPDMC (M, High) & 650.5 & 	6.817e+05 & 	8.75  \\
LSTM-PCPDMC (L, High) & 664.6	 & 7.252e+05 & 	9.00  \\
\hline
\end{tabular}
\end{adjustbox}
\end{center}
\label{tab:el_general_res_selected}
\end{table}

\item \textit{\textbf{RQ4:} Is there an advantage in employing forecast ensembling (MCPDMC-WA) or model switching (MCPDMC-SW) in proximity to change points, with the main consideration being the Symmetric Mean Absolute Percentage Error (SMAPE) metric?}

The performance of the schemas for Mixed Change Point-Divided Model Collection (HT-MCPDMC-WA and HT-MCPDMC-SW) depended on the density of change points (i.e., number of segments). In the case of natural gas consumption forecasting and \textit{Low} PELT settings, the WAVG procedure performed similarly to the HT-PCPDMC approach and slightly reduced the forecast error for all metrics (precise results can be found in Table~\ref{sec:appxB}~\ref{tab:el_general_res_low} and Figure~\ref{fig:sape_low}). However, with more change points, the error reduction was only observed in the first few years of the data, as shown in Figures~\ref{fig:sape_med} and~\ref{fig:sape_high}. After 2015, the HT-PCPDMC approach had sub~13\% SMAPE, while WAVG and SWITCH procedures had SMAPE over 15~\% and failed to beat the baselines. The HT-PCPDMC approach also showed a decreasing trend in SMAPE during incremental learning, similar to the baseline models, but with a lower SMAPE for all PELT settings. Therefore, the HT-PCPDMC approach was more suitable for CL tasks than the HT-MCPDMC approach, as it was less sensitive to the number of detected change points and achieved better results with a simpler procedure as the HT-PCPDMC approach is also not affected by the width of the window near the change point or sudden changes in the errors which may lead to overfitting when the ensembling or switching procedure is employed. Furthermore, the HT-PCPDMC approach had the lowest final SMAPE (year 2020) compared to all other approaches for each PELT setting, as seen in Figures~\ref{fig:sape_low} to~\ref{fig:sape_high}.

We can see the same effect in the electricity load forecasting as in neither PELT setting the WAVG or SWITCH procedures provided lower forecasting error compared to the HT-PCPDMC approach (see Table \ref{tab:el_general_res_selected} and  Tables~\ref{sec:appxB}~\ref{tab:el_general_res_low}~and~\ref{tab:el_general_res_high} or Figures~\ref{fig:sape_low_el}~and~\ref{fig:sape_high_el} for details).

\end{itemize}


\section{Conclusions and future works}
\label{Conclusions}
This article proposed and evaluated two novel approaches to forecasting natural gas consumption based on change point detection and model aggregation. We designed a general methodology for multistep forecasting of multivariate time series with CL capabilities using data stream processing. The methodology consists of several steps: dataset preprocessing, change point detection, model collection setup, model collection selection schema, and model training. We conducted a comparative analysis of the forecast accuracy of the proposed methodologies for model selection employing change point detection using not only natural gas consumption data but also electricity load to ensure applicability of the proposed approaches to other domains as well. We conducted a comparative analysis of the forecast accuracy of the models following the proposed methodologies for model selection employing change point detection with baseline techniques devoid of change point detection and also several deep learning models.
%
We used three metrics (MSE, MAE, and SMAPE), the Diebold-Mariano test, and three settings of the PELT algorithm to measure the forecasting error and the effect of the number of change points. Our experiments showed that the pure change point divided model collection approach outperformed all other approaches by all metrics and for all settings. Deep learning models did not behave well, showing higher errors across various architectures, with little benefit from employing multiple models or change point selection. These findings suggest that Hoeffding Tree-based models, especially those using the proposed HT-PCPDMC scheme, are more effective in forecasting natural gas consumption or electricity load than deep learning approaches. We also found that fewer change points resulted in a lower forecasting error and that the pure approach was more robust and suitable for CL tasks than the mixed approach. Our results demonstrate the potential for using change point detection and model aggregation to forecast natural gas consumption in a dynamic and complex environment.

Using the Hoeffding Tree Regressor for multistep time series forecasting in data streams offers several advantages, particularly when data arrives continuously and in large volumes. However, certain factors must be considered before applying this approach.
First, if the dataset lacks sufficient data or exhibits significant data shifts, it may be more appropriate to utilize offline learning techniques. Offline learning allows the utilization of the entire dataset, enabling more comprehensive model training and evaluation. In contrast, in data stream scenarios with limited data availability, the Hoeffding Tree regressor may not perform optimally due to insufficient data for model training and adaptation.
Moreover, employing the Pruned Exact Linear Time (PELT) algorithm for change point detection in time series data streams is beneficial when the time series contains data shifts or structural changes. In cases where the time series exhibits stability and does not undergo significant shifts, using PELT may not be necessary. However, when data shifts are present, employing PELT can improve forecasting accuracy by identifying and adapting to these changes in the data distribution.
Comparatively, utilizing the Hoeffding Tree regression with PELT change point detection often yields superior results compared to deep learning approaches in scenarios characterized by data shifts. The adaptability of the Hoeffding Tree Regressor to evolving data distributions, coupled with the ability of PELT to detect and respond to changes, contributes to its effectiveness in forecasting time series data streams.

We believe this is an important and promising direction for future research, as it could improve the accuracy and efficiency of forecasting systems and enable them to adapt to changing conditions. Some possible future work includes exploring different types of models and selection schemes, applying the methodology to other domains and datasets, and incorporating uncertainty estimation and anomaly detection into the forecasting pipeline.




\clearpage

\section*{Acknowledgment}
This work was supported by the CEUS-UNISONO programme, which has received funding from the National Science Center, Poland under grant agreement No. 2020/02/Y/ST6/00037, and the GACR-Czech Science Foundation project No. 21-33574K "Lifelong Machine Learning on Data Streams".

\appendix
\section{Features included in the dataset used during the experiments} \label{sec:appxA}
\begin{table}[!ht]
\caption{Description of the features included in the natural gas  dataset.}
\begin{adjustbox}{width=0.6\textwidth,center}
\begin{tabular}{|l|p{9cm}|l|}
\hline
 \multicolumn{1}{|c|}{\textbf{Feature name}} &  \multicolumn{1}{|c|}{\textbf{Description}}                                                                            &  \multicolumn{1}{|c|}{\textbf{Measuring unit}} \\
\hline
Year                  & 2013-2018                                                                                           & -                    \\
\hline
Month                 & 1-12                                                                                                & -                    \\
\hline
Day                   & 1-31                                                                                                & -                  \\
\hline
Hour                  & 0-23                                                                                                & -                 \\
\hline
Day of week           & 1-7                                                                                                 & -                  \\
\hline
Before holiday        & Is the next day a holiday?                                                                            & -                   \\
\hline
Holiday               & Is the current day a holiday?                                                                         & -                     \\
\hline
\multirow{2}*{Temperature}           & Air temperature at a 2-meter height   & 
\multirow{2}*{\si{\celsius}} \\
& above the earth's surface   & \\ 
\hline
Pressure              & Atmospheric pressure at weather station level                                                       &  \si{\milli\meter} Hg \\
\hline
Pressure (sea level)             & Atmospheric pressure reduced to mean sea level                                           &  mm Hg \\
\hline
\multirow{2}*{Humidity}  & Relative humidity at height of 2 meters above the  &  \multirow{2}*{\%}  \\
 & earth's surface  & \\                                 
\hline
\multirow{2}*{Wind direction} & Mean wind direction at a height of $10-12$ meters & \multirow{2}*{compass points} \\
& above  the earth's surface over a 10-min period & \\
\hline
\multirow{2}*{Wind speed}& Mean wind speed at a height of $10-12$ meters   & \multirow{2}*{m/s}      \\
& above the earth's surface over a 10-min period   & \\
\hline
\multirow{2}*{Phenomena}       & Special present weather phenomena observed at & \multirow{2}*{Text value}            \\
& or near the aerodrome & \\
\hline
\multirow{2}*{Recent phenomena}   & Recent weather phenomena of operational  & \multirow{2}*{Text value}     \\
 &  significance                                                & \\ 
\hline
Visibility            & Horizontal visibility                                                                               & km                   \\
\hline
\multirow{2}*{Dewpoint}              & Dewpoint temperature at a height of 2 meters&  \multirow{2}*{\si{\celsius}}     \\
&  above the earth's surface                              & \\
\hline
Cloud cover           & Total cloud cover                                                                                   & Text value         \\
\hline
Price                 & Weighed average                                                                                     & EUR / MWh           \\
\hline
 \multirow{2}*{Consumption} & Consumption of the distribution network     &                             \multirow{2}*{m\textsuperscript{3}/h} \\
 & for the current hour    &\\
 
\hline
Temperature YRNO           & Forecasted temperature       &                                                                  \si{\celsius} \\
\hline
\end{tabular}
\end{adjustbox}
\label{tab:features}
\end{table}

\section{Experiments results in detail}
\label{sec:appxB}

\begin{table}[htp]
\caption{Symmetric Absolute Percentage Error (SAPE) medians of the selected forecasting approaches aggregated by months for 2020 of the natural gas consumption dataset. The lowest median error for the month is in bold.}
\begin{center}
\begin{adjustbox}{width=1.0\textwidth,center}
\begin{tabular}{|l|r|r|r|r|}
\hline
\textbf{Month/Approach} &  HT-PCPDMC (Low)	 & HT-MCPDMC-WA (High) & HT-MCPDMC-SW (High) & FCN-SMCA (M) \\
\hline
January & \textbf{5.53} & 10.14 & 8.61 & 7.62  \\ 
\hline
February & \textbf{6.10} & 6.96 & 7.27  & 8.31  \\ 
\hline
March & \textbf{7.78} & 14.29 & 14.22  &  8.97 \\ 
\hline
April & \textbf{14.74} & 30.66 & 25.76 & 16.49  \\ 
\hline
May & \textbf{14.69} & 15.27 & 15.27 &  17.60 \\ 
\hline
June & 6.91 & 6.91 & \textbf{7.53}  & 9.23 \\ 
\hline
July & \textbf{8.20} & 8.34 & 8.34  & 15.95 \\ 
\hline
August & \textbf{7.33} & 7.99 & 7.99  & 13.40 \\ 
\hline
September & 9.98 & 9.42 & \textbf{8.42}  & 11.05 \\ 
\hline
October & 10.55 & \textbf{9.57} & 10.27  & 9.61 \\ 
\hline
November & 8.20 & 7.98 & 8.79  & \textbf{7.85} \\ 
\hline
December & \textbf{5.66} & 8.14 & 7.48 & 6.82 \\
\hline
\end{tabular}
\end{adjustbox}
\end{center}

\label{tab:sape_by_month}
\end{table}

\begin{table}[htp]
\caption{Symmetric Absolute Percentage Error (SAPE) medians of the selected forecasting approaches aggregated by months for 2020 of the electricity load dataset. The lowest median error for the month is in bold.}
\begin{center}
\begin{adjustbox}{width=1.0\textwidth,center}
\begin{tabular}{|l|r|r|r|r|}
\hline
\textbf{Month/Approach} &  HT-PCPDMC (Low)	 & HT-MCPDMC-WA (High) & HT-MCPDMC-SW (High) & FCN-SMCA (S) \\
\hline
January & 4.01 & 	4.43 & 	\textbf{3.37} &	8.91  \\ 
\hline
February & 5.39	 & 4.96 & 	\textbf{3.48} &	7.10  \\ 
\hline
March & 6.32 & 	4.90	 & \textbf{4.32} &	4.88 \\ 
\hline
April & \textbf{6.70} & 	8.47 & 	8.03 &	7.04  \\ 
\hline
May & 6.88 & 	7.52 & 	7.52 &	\textbf{6.14} \\ 
\hline
June & \textbf{6.57} & 	8.54 & 	8.54 &	7.67 \\ 
\hline
July & 5.98 & 	5.09 & 	\textbf{4.54} &	6.29 \\ 
\hline
August & 4.88 & 	4.85 & 	\textbf{4.66} &	5.92 \\ 
\hline
September & \textbf{5.15} & 	5.47 & 	5.47 &	7.55 \\ 
\hline
October & 6.16 & 	\textbf{5.68} & 	\textbf{5.68} &	7.16 \\ 
\hline
November & \textbf{4.59}	 & 5.99 & 	5.57 &	8.20 \\ 
\hline
December & \textbf{6.83} & 	9.75 & 	8.64 &	7.40 \\
\hline
\end{tabular}
\end{adjustbox}
\end{center}
\label{tab:sape_by_month_el}
\end{table}

\begin{table}[!htp]
\caption{Summarized results of experiments conducted over natural gas consumption dataset using the baseline approaches. The MAE, MSE and SMAPE measures were computed for the forecasted data from January 1, 2014, to December 31, 2020.}
\begin{center}
\begin{adjustbox}{width=0.75\textwidth,center}
\begin{tabular}{|l|r|r|r|}
\hline
\textbf{Approach} & \textbf{MAE} & \textbf{MSE} & \textbf{SMAPE} \\
\hline
HT-SMCA              & 1.170e+04     & 3.143e+08    & \textbf{12.94}          \\
HT-QDMDC             & 1.179e+04    & 3.109e+08    & 13.14          \\
\hline
FCN-SMCA (S) & 1.588e+04 &	5.866e+08 &	16.70 \\
FCN-SMCA (M) & 1.548e+04 & 5.398e+08 &	16.49 \\
FCN-SMCA (L) & 1.655e+04 & 6.052e+08 & 17.09 \\
GRU-SMCA (S) & 1.936e+04 &	8.117e+08 &	20.68 \\
GRU-SMCA (M) & 1.919e+04 &	7.970e+08 &	20.26 \\
GRU-SMCA (L) & 1.798e+04 &	7.080e+08 &	19.03 \\
LSTM-SMCA (S) & 1.948e+04 &	7.924e+08 &	21.71 \\
LSTM-SMCA (M) & 1.915e+04 &	8.085e+08 &	21.65 \\
LSTM-SMCA (L) & 1.879e+04 &	7.557e+08 &	20.58 \\
\hline
FCN-QDMDC (S) & 1.983e+04  & 	8.698e+08  & 	21.61 \\
FCN-QDMDC (M) & 1.938e+04  & 	8.493e+08  & 	21.08 \\
FCN-QDMDC (L) & 2.100e+04  & 	9.381e+08 	 & 25.77 \\
GRU-QDMDC (S) & 2.605e+04  & 	1.497e+09  & 	30.06 \\
GRU-QDMDC (M) & 2.306e+04  & 	1.109e+09  & 	27.50 \\
GRU-QDMDC (L) & 2.352e+04  & 	1.121e+09  & 	31.46 \\
LSTM-QDMDC (S) & 2.612e+04  & 	1.401e+09  & 	31.10 \\
LSTM-QDMDC (M) & 3.301e+04	 & 3.058e+09  & 	34.81 \\
LSTM-QDMDC (L) & 3.582e+04  & 	3.114e+09	  & 50.19 \\
\hline
\end{tabular}
\end{adjustbox}
\end{center}
\label{tab:gas_general_res}
\end{table}

\begin{table}[!htp]
\caption{Summarized results of experiments conducted over natural gas consumption dataset using the proposed approaches for the \textit{Low} number of change points. The MAE, MSE and SMAPE measures were computed for the forecasted data from January 1, 2014, to December 31, 2020.}
\begin{center}
\begin{tabular}{|l|r|r|r|}
\hline
\textbf{Approach} & \textbf{MAE} & \textbf{MSE} & \textbf{SMAPE} \\
\hline
HT-PCPDMC            & 1.110e+04     & 2.783e+08    & 12.32          \\
HT-MCPDMC-WA       & 1.108e+04    & 2.770e+08     & \textbf{12.29}          \\
HT-MCPDMC-SW     & 1.124e+04    & 2.856e+08    & 12.48          \\
\hline
FCN-PCPDMC (S) & 1.902e+04 &	7.643e+08 &	21.41 \\
FCN-PCPDMC (M) & 1.750e+04 &	6.457e+08 &	20.34  \\
FCN-PCPDMC (L) & 1.780e+04 &	6.650e+08 &	21.49 \\
\hline
GRU-PCPDMC (S) & 2.325e+04 &	1.115e+09 &	28.11  \\
GRU-PCPDMC (M) & 2.185e+04 &	1.003e+09 &	24.89 \\
GRU-PCPDMC (L) & 2.107e+04 &	9.241e+08 &	24.55  \\
\hline
LSTM-PCPDMC (S) & 2.786e+04 &	1.781e+09 &	45.80  \\
LSTM-PCPDMC (M) & 2.533e+04 &	1.081e+09 &	44.68  \\
LSTM-PCPDMC (L) & 2.143e+04 &	9.764e+08 &	24.23  \\
\hline
\end{tabular}
\end{center}

\label{tab:gas_general_res_low}
\end{table}

\begin{table}[!htp]
\caption{Summarized results of experiments conducted over natural gas consumption dataset using the proposed approaches for the \textit{Medium} number of change points. The MAE, MSE and SMAPE measures were computed for the forecasted data from January 1, 2014, to December 31, 2020.}
\begin{center}
\begin{tabular}{|l|r|r|r|}
\hline
\textbf{Approach} & \textbf{MAE} & \textbf{MSE} & \textbf{SMAPE} \\
\hline
HT-PCPDMC        & 1.181e+04    & 3.211e+08    & \textbf{12.60}           \\
HT-MCPDMC-WA     & 1.196e+04    & 3.233e+08    & 12.78          \\
HT-MCPDMC-SW     & 1.189e+04    & 3.223e+08    & 12.81          \\
\hline
FCN-PCPDMC (S) & 2.090e+04 & 	9.454e+08 & 	22.43 \\
FCN-PCPDMC (M) & 2.030e+04 & 	8.837e+08 & 	22.17 \\
FCN-PCPDMC (L) & 2.010e+04 & 	8.929e+08	 & 22.08 \\
\hline
GRU-PCPDMC (S) & 2.774e+04 & 	1.331e+09 & 	45.60  \\
GRU-PCPDMC (M) & 2.299e+04	 & 1.105e+09 & 	26.51 \\
GRU-PCPDMC (L) & 2.270e+04	 & 1.091e+09 & 	26.54  \\
\hline
LSTM-PCPDMC (S) & 4.856e+04	 & 8.031e+09 & 	51.02  \\
LSTM-PCPDMC (M) & 3.275e+04 & 	3.208e+09 & 	45.85  \\
LSTM-PCPDMC (L) & 2.187e+04 & 	1.053e+09 & 	23.87  \\
\hline
\end{tabular}
\end{center}

\label{tab:gas_general_res_med}
\end{table}

\begin{table}[!htp]
\caption{Summarized results of experiments conducted over natural gas consumption dataset using the proposed approaches for the \textit{High} number of change points. The MAE, MSE and SMAPE measures were computed for the forecasted data from January 1, 2014, to December 31, 2020.}
\begin{center}
\begin{tabular}{|l|r|r|r|}
\hline
\textbf{Approach} & \textbf{MAE} & \textbf{MSE} & \textbf{SMAPE} \\
\hline
HT-PCPDMC            & 1.265e+04    & 3.925e+08    & \textbf{13.09}          \\
HT-MCPDMC-WA       & 1.557e+04    & 6.368e+08    & 15.12          \\
HT-MCPDMC-SW     & 1.483e+04    & 5.711e+08    & 14.72    \\
\hline
FCN-PCPDMC (S) & 2.100e+04 &	9.688e+08 &	22.28 \\
FCN-PCPDMC (M) & 1.975e+04 &	8.508e+08 &	21.05 \\
FCN-PCPDMC (L) & 1.977e+04 &	8.391e+08 &	22.16 \\
\hline
GRU-PCPDMC (S) & 2.450e+04	 &1.265e+09 &	29.71  \\
GRU-PCPDMC (M) & 2.264e+04 &	1.029e+09 &	27.60 \\
GRU-PCPDMC (L) & 2.173e+04 &	9.828e+08 &	25.01  \\
\hline
LSTM-PCPDMC (S) & 2.498e+04 &	1.387e+09 &	29.19  \\
LSTM-PCPDMC (M) & 3.183e+04 &	3.135e+09 &	45.33  \\
LSTM-PCPDMC (L) & 2.097e+04 &	9.366e+08 &	23.11  \\
\hline
\end{tabular}
\end{center}

\label{tab:gas_general_res_high}
\end{table}

\begin{table}[!htp]
\caption{Summarized results of experiments conducted over electricity load dataset using the baseline approaches. The MAE, MSE and SMAPE measures were computed for the forecasted data from January 1, 2014, to December 31, 2020.}
\begin{center}
\begin{adjustbox}{width=0.75\textwidth,center}
\begin{tabular}{|l|r|r|r|}
\hline
\textbf{Approach} & \textbf{MAE} & \textbf{MSE} & \textbf{SMAPE} \\
\hline
HT-SMCA              &      507.2 & 	5.053e+05 & 	\textbf{6.84}     \\
HT-QDMDC             &      537.6 & 	5.480e+05 & 	7.25     \\
\hline
FCN-SMCA (S) & 573.5 & 	5.571e+05 & 	7.73 \\
FCN-SMCA (M) & 592.1 & 	5.906e+05 & 	7.97   \\
FCN-SMCA (L) & 580.7 & 	5.764e+05 & 	7.82   \\
GRU-SMCA (S) & 673.2 & 	7.455e+05 & 	9.10   \\
GRU-SMCA (M) & 642.2 & 	6.975e+05 & 	8.68   \\
GRU-SMCA (L) & 636.1 & 	6.873e+05 & 	8.60   \\
LSTM-SMCA (S) & 683.4	 & 7.664e+05 & 	9.26   \\
LSTM-SMCA (M) & 627.6	 & 6.698e+05 & 	8.49   \\
LSTM-SMCA (L) & 646.3 & 	7.036e+05 & 	8.74   \\
\hline
FCN-QDMDC (S) & 659.5 & 	6.967e+05 & 	8.87   \\
FCN-QDMDC (M) & 646.1	 & 6.733e+05 & 	8.68   \\
FCN-QDMDC (L) & 646.5 & 	6.749e+05 & 	8.69  \\
GRU-QDMDC (S) & 690.8 & 	7.757e+05 & 	9.34  \\
GRU-QDMDC (M) & 668.0 & 	7.338e+05 & 	9.05   \\
GRU-QDMDC (L) & 672.0 & 	7.493e+05 & 	9.09   \\
LSTM-QDMDC (S) & 842.1 & 	1.428e+06 & 	11.45  \\
LSTM-QDMDC (M) & 673.8 & 	7.459e+05 & 	9.11  \\
LSTM-QDMDC (L) & 665.4 & 	7.370e+05 & 	9.00  \\
\hline
\end{tabular}
\end{adjustbox}
\end{center}

\label{tab:el_general_res}
\end{table}

\begin{table}[!htp]
\caption{Summarized results of experiments conducted over electricity load dataset using the proposed approaches for the \textit{Low} number of change points. The MAE, MSE and SMAPE measures were computed for the forecasted data from January 1, 2014, to December 31, 2020.}
\begin{center}
\begin{tabular}{|l|r|r|r|}
\hline
\textbf{Approach} & \textbf{MAE} & \textbf{MSE} & \textbf{SMAPE} \\
\hline
HT-PCPDMC            & 506.5 & 	5.020e+05 & 	\textbf{6.83}         \\
HT-MCPDMC-WA       & 522.6 & 	5.344e+05 & 	7.05          \\
HT-MCPDMC-SW     & 526.6 & 	5.445e+05 & 	7.10          \\
\hline
FCN-PCPDMC (S) & 624.6 & 	6.380e+05 & 	8.38 \\
FCN-PCPDMC (M) & 611.6 & 	6.251e+05 & 	8.22  \\
FCN-PCPDMC (L) & 612.2 & 	6.276e+05 & 	8.23 \\
\hline
GRU-PCPDMC (S) & 649.3 & 	7.063e+05 & 	8.77  \\
GRU-PCPDMC (M) & 642.3 & 	6.948e+05 & 	8.68 \\
GRU-PCPDMC (L) & 634.0	 & 6.783e+05 & 	8.57  \\
\hline
LSTM-PCPDMC (S) & 648.1 & 	7.066e+05 & 	8.74  \\
LSTM-PCPDMC (M) & 635.2 & 	6.838e+05 & 	8.58  \\
LSTM-PCPDMC (L) & 633.1 & 	6.768e+05 & 	8.56  \\
\hline
\end{tabular}
\end{center}

\label{tab:el_general_res_low}
\end{table}

\begin{table}[!htp]
\caption{Summarized results of experiments conducted over electricity load dataset using the proposed approaches for the \textit{Medium} number of change points. The MAE, MSE and SMAPE measures were computed for the forecasted data from January 1, 2014, to December 31, 2020.}
\begin{center}
\begin{tabular}{|l|r|r|r|}
\hline
\textbf{Approach} & \textbf{MAE} & \textbf{MSE} & \textbf{SMAPE} \\
\hline
HT-PCPDMC            & 506.2 &	5.001e+05 &	\textbf{6.83}         \\
HT-MCPDMC-WA       & 532.1 &	5.467e+05 &	7.17          \\
HT-MCPDMC-SW     & 533.3 &	5.555e+05 &	7.11         \\
\hline
FCN-PCPDMC (S) & 625.2 &	6.473e+05 &	8.40 \\
FCN-PCPDMC (M) & 616.2 &	6.368e+05 &	8.27  \\
FCN-PCPDMC (L) & 615.1 &	6.361e+05 &	8.26 \\
\hline
GRU-PCPDMC (S) & 649.3 &	7.091e+05 &	8.76  \\
GRU-PCPDMC (M) & 637.8 &	6.847e+05 &	8.63 \\
GRU-PCPDMC (L) & 634.7 &	6.865e+05 &	8.58  \\
\hline
LSTM-PCPDMC (S) & 662.7 &	7.426e+05 &	8.92  \\
LSTM-PCPDMC (M) & 724.7 &	1.002e+06 &	9.55  \\
LSTM-PCPDMC (L) & 625.6 &	6.661e+05 &	8.46  \\
\hline
\end{tabular}
\end{center}

\label{tab:el_general_res_med}
\end{table}

\begin{table}[!htp]
\caption{Summarized results of experiments conducted over electricity load dataset using the proposed approaches for the \textit{High} number of change points. The MAE, MSE and SMAPE measures were computed for the forecasted data from January 1, 2014, to December 31, 2020.}
\begin{center}
\begin{tabular}{|l|r|r|r|}
\hline
\textbf{Approach} & \textbf{MAE} & \textbf{MSE} & \textbf{SMAPE} \\
\hline
HT-PCPDMC            & 517.2 & 	5.197e+05 & 	\textbf{6.99}         \\
HT-MCPDMC-WA       & 576.0 & 	6.168e+05 & 	7.77          \\
HT-MCPDMC-SW     & 573.1 & 	6.278e+05 & 	7.74         \\
\hline
FCN-PCPDMC (S) & 669.2 & 	7.163e+05 & 	9.01 \\
FCN-PCPDMC (M) & 650.5 & 	6.817e+05 & 	8.75  \\
FCN-PCPDMC (L) & 653.1 & 	6.855e+05 & 	8.79 \\
\hline
GRU-PCPDMC (S) & 696.5 & 	7.957e+05 & 	9.42  \\
GRU-PCPDMC (M) & 675.3 & 	7.422e+05 & 	9.12 \\
GRU-PCPDMC (L) & 667.0 & 	7.297e+05 & 	9.01  \\
\hline
LSTM-PCPDMC (S) & 	707.6 & 	8.227e+05 & 	9.55  \\
LSTM-PCPDMC (M) & 674.7 & 	7.422e+05 & 	9.12  \\
LSTM-PCPDMC (L) & 664.6	 & 7.252e+05 & 	9.00  \\
\hline
\end{tabular}
\end{center}

\label{tab:el_general_res_high}
\end{table}

\clearpage
\newpage

\begingroup
\let\clearpage\relax

\setcitestyle{numbers}
\bibliographystyle{elsarticle-num}
\bibliography{refs}

\begin{thebibliography}{10}
\expandafter\ifx\csname url\endcsname\relax
  \def\url#1{\texttt{#1}}\fi
\expandafter\ifx\csname urlprefix\endcsname\relax\def\urlprefix{URL }\fi
\expandafter\ifx\csname href\endcsname\relax
  \def\href#1#2{#2} \def\path#1{#1}\fi

\bibitem{creti2004long}
A.~Creti, B.~Villeneuve, et~al., Long-term contracts and take-or-pay clauses in natural gas markets, Energy Studies Review 13~(1) (2004) 75--94.
\newblock \href {https://doi.org/http://dx.doi.org/10.15173/esr.v13i1.466} {\path{doi:http://dx.doi.org/10.15173/esr.v13i1.466}}.

\bibitem{medina1986take}
J.~M. Medina, G.~A. McKenzie, B.~M. Daniel, \href{https://www.academia.edu/2446956/Take_or_Litigate_Enforcing_the_Plain_Meaning_of_the_Take_or_Pay_Clause_in_Natural_Gas_Contracts}{Take or litigate: Enforcing the plain meaning of the take-or-pay clause in natural gas contracts}, Ark. L. Rev. 40 (1986) 185.
\newline\urlprefix\url{https://www.academia.edu/2446956/Take_or_Litigate_Enforcing_the_Plain_Meaning_of_the_Take_or_Pay_Clause_in_Natural_Gas_Contracts}

\bibitem{Balestra1966}
P.~Balestra, M.~Nerlove, \href{http://www.jstor.org/stable/1909771}{Pooling cross section and time series data in the estimation of a dynamic model: The demand for natural gas}, Econometrica 34~(3) (1966) 585--612.
\newline\urlprefix\url{http://www.jstor.org/stable/1909771}

\bibitem{Vondracek2008}
J.~Vondráček, E.~Pelikán, O.~Konár, J.~Čermáková, K.~Eben, M.~Malý, M.~Brabec, \href{http://www.sciencedirect.com/science/article/pii/S0306261907001183}{A statistical model for the estimation of natural gas consumption}, Applied Energy 85~(5) (2008) 362 -- 370.
\newblock \href {https://doi.org/https://doi.org/10.1016/j.apenergy.2007.07.004} {\path{doi:https://doi.org/10.1016/j.apenergy.2007.07.004}}.
\newline\urlprefix\url{http://www.sciencedirect.com/science/article/pii/S0306261907001183}

\bibitem{Suykens1996}
J.~Suykens, P.~Lemmerling, W.~Favoreel, B.~De~Moor, M.~Crepel, P.~Briol, Modelling the belgian gas consumption using neural networks, Neural processing letters 4~(3) (1996) 157--166.
\newblock \href {https://doi.org/https://doi.org/10.1007/BF00426024} {\path{doi:https://doi.org/10.1007/BF00426024}}.

\bibitem{Khotanzad2000}
A.~{Khotanzad}, H.~{Elragal}, T.~. {Lu}, Combination of artificial neural-network forecasters for prediction of natural gas consumption, IEEE Transactions on Neural Networks 11~(2) (2000) 464--473.
\newblock \href {https://doi.org/https://doi.org/10.1109/72.839015} {\path{doi:https://doi.org/10.1109/72.839015}}.

\bibitem{Viet2005}
N.~H. Viet, J.~Mańdziuk, Neural and fuzzy neural networks in prediction of natural gas consumption, Neural Parallel \& Scientific Comp. 13 (2005) 265--286.
\newblock \href {https://doi.org/https://doi.org/10.1109/NNSP.2003.1318075} {\path{doi:https://doi.org/10.1109/NNSP.2003.1318075}}.

\bibitem{Soldo2014}
B.~Soldo, P.~Potočnik, G.~Šimunović, T.~Šarić, E.~Govekar, \href{http://www.sciencedirect.com/science/article/pii/S0378778813007299}{Improving the residential natural gas consumption forecasting models by using solar radiation}, Energy and Buildings 69 (2014) 498 -- 506.
\newblock \href {https://doi.org/https://doi.org/10.1016/j.enbuild.2013.11.032} {\path{doi:https://doi.org/10.1016/j.enbuild.2013.11.032}}.
\newline\urlprefix\url{http://www.sciencedirect.com/science/article/pii/S0378778813007299}

\bibitem{Taspinar2013}
F.~Taşpınar, N.~Çelebi, N.~Tutkun, \href{http://www.sciencedirect.com/science/article/pii/S0378778812005324}{Forecasting of daily natural gas consumption on regional basis in turkey using various computational methods}, Energy and Buildings 56 (2013) 23 -- 31.
\newblock \href {https://doi.org/https://doi.org/10.1016/j.enbuild.2012.10.023} {\path{doi:https://doi.org/10.1016/j.enbuild.2012.10.023}}.
\newline\urlprefix\url{http://www.sciencedirect.com/science/article/pii/S0378778812005324}

\bibitem{Boran2015}
F.~E. Boran, \href{https://doi.org/10.1080/15567249.2014.893040}{Forecasting natural gas consumption in turkey using grey prediction}, Energy Sources, Part B: Economics, Planning, and Policy 10~(2) (2015) 208--213.
\newblock \href {http://arxiv.org/abs/https://doi.org/10.1080/15567249.2014.893040} {\path{arXiv:https://doi.org/10.1080/15567249.2014.893040}}, \href {https://doi.org/10.1080/15567249.2014.893040} {\path{doi:10.1080/15567249.2014.893040}}.
\newline\urlprefix\url{https://doi.org/10.1080/15567249.2014.893040}

\bibitem{Szoplik2015}
J.~Szoplik, \href{http://www.sciencedirect.com/science/article/pii/S036054421500393X}{Forecasting of natural gas consumption with artificial neural networks}, Energy 85 (2015) 208 -- 220.
\newblock \href {https://doi.org/https://doi.org/10.1016/j.energy.2015.03.084} {\path{doi:https://doi.org/10.1016/j.energy.2015.03.084}}.
\newline\urlprefix\url{http://www.sciencedirect.com/science/article/pii/S036054421500393X}

\bibitem{rahman2017}
A.~Rahman, A.~D. Smith, Predicting fuel consumption for commercial buildings with machine learning algorithms, Energy and Buildings 152 (2017) 341 -- 358.
\newblock \href {https://doi.org/https://doi.org/10.1016/j.enbuild.2017.07.017} {\path{doi:https://doi.org/10.1016/j.enbuild.2017.07.017}}.

\bibitem{Su2019}
H.~Su, E.~Zio, J.~Zhang, M.~Xu, X.~Li, Z.~Zhang, A hybrid hourly natural gas demand forecasting method based on the integration of wavelet transform and enhanced deep-rnn model, Energy 178 (2019) 585 -- 597.
\newblock \href {https://doi.org/https://doi.org/10.1016/j.energy.2019.04.167} {\path{doi:https://doi.org/10.1016/j.energy.2019.04.167}}.

\bibitem{WEI2023205133}
N.~Wei, L.~Yin, C.~Yin, J.~Liu, S.~Wang, W.~Qiao, F.~Zeng, \href{https://www.sciencedirect.com/science/article/pii/S2949908923002613}{Pseudo-correlation problem and its solution for the transfer forecasting of short-term natural gas loads}, Gas Science and Engineering 119 (2023) 205133.
\newblock \href {https://doi.org/https://doi.org/10.1016/j.jgsce.2023.205133} {\path{doi:https://doi.org/10.1016/j.jgsce.2023.205133}}.
\newline\urlprefix\url{https://www.sciencedirect.com/science/article/pii/S2949908923002613}

\bibitem{WEI2024122087}
N.~Wei, C.~Yin, L.~Yin, J.~Tan, J.~Liu, S.~Wang, W.~Qiao, F.~Zeng, \href{https://www.sciencedirect.com/science/article/pii/S0306261923014514}{Short-term load forecasting based on wm algorithm and transfer learning model}, Applied Energy 353 (2024) 122087.
\newblock \href {https://doi.org/https://doi.org/10.1016/j.apenergy.2023.122087} {\path{doi:https://doi.org/10.1016/j.apenergy.2023.122087}}.
\newline\urlprefix\url{https://www.sciencedirect.com/science/article/pii/S0306261923014514}

\bibitem{BAI2016}
Y.~Bai, C.~Li, \href{http://www.sciencedirect.com/science/article/pii/S0378778816305096}{Daily natural gas consumption forecasting based on a structure-calibrated support vector regression approach}, Energy and Buildings 127 (2016) 571 -- 579.
\newblock \href {https://doi.org/https://doi.org/10.1016/j.enbuild.2016.06.020} {\path{doi:https://doi.org/10.1016/j.enbuild.2016.06.020}}.
\newline\urlprefix\url{http://www.sciencedirect.com/science/article/pii/S0378778816305096}

\bibitem{SOLDO2012}
B.~Soldo, Forecasting natural gas consumption, Applied Energy 92 (2012) 26 -- 37.
\newblock \href {https://doi.org/https://doi.org/10.1016/j.apenergy.2011.11.003} {\path{doi:https://doi.org/10.1016/j.apenergy.2011.11.003}}.

\bibitem{Tamba2018}
J.~G. Tamba, S.~N. Essiane, E.~F. Sapnken, F.~D. Koffi, J.~L. Nsouandélé, B.~Soldo, D.~Njomo, \href{https://ideas.repec.org/a/eco/journ2/2018-03-28.html}{Forecasting natural gas: A literature survey}, International Journal of Energy Economics and Policy 8~(3) (2018) 216--249.
\newline\urlprefix\url{https://ideas.repec.org/a/eco/journ2/2018-03-28.html}

\bibitem{Chen:2018}
Z.~Chen, B.~Liu, R.~Brachman, P.~Stone, F.~Rossi, Lifelong Machine Learning, 2nd Edition, Morgan {\&} Claypool Publishers, 2018.

\bibitem{FRENCH1999128}
R.~M. French, \href{https://www.sciencedirect.com/science/article/pii/S1364661399012942}{Catastrophic forgetting in connectionist networks}, Trends in Cognitive Sciences 3~(4) (1999) 128--135.
\newblock \href {https://doi.org/https://doi.org/10.1016/S1364-6613(99)01294-2} {\path{doi:https://doi.org/10.1016/S1364-6613(99)01294-2}}.
\newline\urlprefix\url{https://www.sciencedirect.com/science/article/pii/S1364661399012942}

\bibitem{Knoblauch2020}
J.~Knoblauch, H.~Husain, T.~Diethe, \href{https://www.amazon.science/publications/optimal-continual-learning-has-perfect-memory-and-is-np-hard}{Optimal continual learning has perfect memory and is np-hard}, in: ICML 2020, 2020.
\newline\urlprefix\url{https://www.amazon.science/publications/optimal-continual-learning-has-perfect-memory-and-is-np-hard}

\bibitem{PARISI201954}
G.~I. Parisi, R.~Kemker, J.~L. Part, C.~Kanan, S.~Wermter, \href{https://www.sciencedirect.com/science/article/pii/S0893608019300231}{Continual lifelong learning with neural networks: A review}, Neural Networks 113 (2019) 54--71.
\newblock \href {https://doi.org/https://doi.org/10.1016/j.neunet.2019.01.012} {\path{doi:https://doi.org/10.1016/j.neunet.2019.01.012}}.
\newline\urlprefix\url{https://www.sciencedirect.com/science/article/pii/S0893608019300231}

\bibitem{DBLP:journals/corr/abs-1902-10486}
A.~Chaudhry, M.~Rohrbach, M.~Elhoseiny, T.~Ajanthan, P.~K. Dokania, P.~H.~S. Torr, M.~Ranzato, \href{http://arxiv.org/abs/1902.10486}{Continual learning with tiny episodic memories}, CoRR abs/1902.10486 (2019).
\newblock \href {http://arxiv.org/abs/1902.10486} {\path{arXiv:1902.10486}}.
\newline\urlprefix\url{http://arxiv.org/abs/1902.10486}

\bibitem{NEURIPS2019_15825aee}
R.~Aljundi, E.~Belilovsky, T.~Tuytelaars, L.~Charlin, M.~Caccia, M.~Lin, L.~Page-Caccia, \href{https://proceedings.neurips.cc/paper_files/paper/2019/file/15825aee15eb335cc13f9b559f166ee8-Paper.pdf}{Online continual learning with maximal interfered retrieval}, in: H.~Wallach, H.~Larochelle, A.~Beygelzimer, F.~d\textquotesingle Alch\'{e}-Buc, E.~Fox, R.~Garnett (Eds.), Advances in Neural Information Processing Systems, Vol.~32, Curran Associates, Inc., 2019.
\newline\urlprefix\url{https://proceedings.neurips.cc/paper_files/paper/2019/file/15825aee15eb335cc13f9b559f166ee8-Paper.pdf}

\bibitem{caccia2022new}
L.~Caccia, R.~Aljundi, N.~Asadi, T.~Tuytelaars, J.~Pineau, E.~Belilovsky, \href{https://openreview.net/forum?id=N8MaByOzUfb}{New insights on reducing abrupt representation change in online continual learning}, in: International Conference on Learning Representations, 2022.
\newline\urlprefix\url{https://openreview.net/forum?id=N8MaByOzUfb}

\bibitem{NEURIPS2022_5ebbbac6}
Y.~Zhang, B.~Pfahringer, E.~Frank, A.~Bifet, N.~J.~S. Lim, Y.~Jia, \href{https://proceedings.neurips.cc/paper_files/paper/2022/file/5ebbbac62b968254093023f1c95015d3-Paper-Conference.pdf}{A simple but strong baseline for online continual learning: Repeated augmented rehearsal}, in: S.~Koyejo, S.~Mohamed, A.~Agarwal, D.~Belgrave, K.~Cho, A.~Oh (Eds.), Advances in Neural Information Processing Systems, Vol.~35, Curran Associates, Inc., 2022, pp. 14771--14783.
\newline\urlprefix\url{https://proceedings.neurips.cc/paper_files/paper/2022/file/5ebbbac62b968254093023f1c95015d3-Paper-Conference.pdf}

\bibitem{NEURIPS2020_b704ea2c}
P.~Buzzega, M.~Boschini, A.~Porrello, D.~Abati, S.~CALDERARA, \href{https://proceedings.neurips.cc/paper_files/paper/2020/file/b704ea2c39778f07c617f6b7ce480e9e-Paper.pdf}{Dark experience for general continual learning: a strong, simple baseline}, in: H.~Larochelle, M.~Ranzato, R.~Hadsell, M.~Balcan, H.~Lin (Eds.), Advances in Neural Information Processing Systems, Vol.~33, Curran Associates, Inc., 2020, pp. 15920--15930.
\newline\urlprefix\url{https://proceedings.neurips.cc/paper_files/paper/2020/file/b704ea2c39778f07c617f6b7ce480e9e-Paper.pdf}

\bibitem{9891836}
M.~Boschini, L.~Bonicelli, P.~Buzzega, A.~Porrello, S.~Calderara, Class-incremental continual learning into the extended der-verse, IEEE Transactions on Pattern Analysis and Machine Intelligence 45~(5) (2023) 5497--5512.
\newblock \href {https://doi.org/10.1109/TPAMI.2022.3206549} {\path{doi:10.1109/TPAMI.2022.3206549}}.

\bibitem{doi:10.1073/pnas.1611835114}
J.~Kirkpatrick, R.~Pascanu, N.~Rabinowitz, J.~Veness, G.~Desjardins, A.~A. Rusu, K.~Milan, J.~Quan, T.~Ramalho, A.~Grabska-Barwinska, D.~Hassabis, C.~Clopath, D.~Kumaran, R.~Hadsell, \href{https://www.pnas.org/doi/abs/10.1073/pnas.1611835114}{Overcoming catastrophic forgetting in neural networks}, Proceedings of the National Academy of Sciences 114~(13) (2017) 3521--3526.
\newblock \href {http://arxiv.org/abs/https://www.pnas.org/doi/pdf/10.1073/pnas.1611835114} {\path{arXiv:https://www.pnas.org/doi/pdf/10.1073/pnas.1611835114}}, \href {https://doi.org/10.1073/pnas.1611835114} {\path{doi:10.1073/pnas.1611835114}}.
\newline\urlprefix\url{https://www.pnas.org/doi/abs/10.1073/pnas.1611835114}

\bibitem{si}
F.~Zenke, B.~Poole, S.~Ganguli, Continual learning through synaptic intelligence, Proceedings of machine learning research 70 (2017) 3987–3995.
\newblock \href {https://doi.org/https://doi.org/10.48550/arXiv.1703.04200} {\path{doi:https://doi.org/10.48550/arXiv.1703.04200}}.

\bibitem{DBLP:journals/corr/RusuRDSKKPH16}
A.~A. Rusu, N.~C. Rabinowitz, G.~Desjardins, H.~Soyer, J.~Kirkpatrick, K.~Kavukcuoglu, R.~Pascanu, R.~Hadsell, \href{http://arxiv.org/abs/1606.04671}{Progressive neural networks}, CoRR abs/1606.04671 (2016).
\newblock \href {http://arxiv.org/abs/1606.04671} {\path{arXiv:1606.04671}}.
\newline\urlprefix\url{http://arxiv.org/abs/1606.04671}

\bibitem{ermis2022memory}
B.~Ermis, G.~Zappella, M.~Wistuba, A.~Rawal, C.~Archambeau, \href{https://openreview.net/forum?id=U07d1Y-x2E}{Memory efficient continual learning with transformers}, in: A.~H. Oh, A.~Agarwal, D.~Belgrave, K.~Cho (Eds.), Advances in Neural Information Processing Systems, 2022.
\newline\urlprefix\url{https://openreview.net/forum?id=U07d1Y-x2E}

\bibitem{9679108}
V.~Gupta, J.~Narwariya, P.~Malhotra, L.~Vig, G.~Shroff, Continual learning for multivariate time series tasks with variable input dimensions, in: 2021 IEEE International Conference on Data Mining (ICDM), 2021, pp. 161--170.
\newblock \href {https://doi.org/10.1109/ICDM51629.2021.00026} {\path{doi:10.1109/ICDM51629.2021.00026}}.

\bibitem{10.1145/3517745.3563033}
G.~G. Gonz\'{a}lez, P.~Casas, A.~Fern\'{a}ndez, G.~G\'{o}mez, \href{https://doi.org/10.1145/3517745.3563033}{Steps towards continual learning in multivariate time-series anomaly detection using variational autoencoders}, in: Proceedings of the 22nd ACM Internet Measurement Conference, IMC '22, Association for Computing Machinery, New York, NY, USA, 2022, p. 774–775.
\newblock \href {https://doi.org/10.1145/3517745.3563033} {\path{doi:10.1145/3517745.3563033}}.
\newline\urlprefix\url{https://doi.org/10.1145/3517745.3563033}

\bibitem{COSSU2021607}
A.~Cossu, A.~Carta, V.~Lomonaco, D.~Bacciu, \href{https://www.sciencedirect.com/science/article/pii/S0893608021002847}{Continual learning for recurrent neural networks: An empirical evaluation}, Neural Networks 143 (2021) 607--627.
\newblock \href {https://doi.org/https://doi.org/10.1016/j.neunet.2021.07.021} {\path{doi:https://doi.org/10.1016/j.neunet.2021.07.021}}.
\newline\urlprefix\url{https://www.sciencedirect.com/science/article/pii/S0893608021002847}

\bibitem{TONG2023106005}
M.~Tong, F.~Qin, J.~Dong, \href{https://www.sciencedirect.com/science/article/pii/S0952197623001896}{Natural gas consumption forecasting using an optimized grey bernoulli model: The case of the world’s top three natural gas consumers}, Engineering Applications of Artificial Intelligence 122 (2023) 106005.
\newblock \href {https://doi.org/https://doi.org/10.1016/j.engappai.2023.106005} {\path{doi:https://doi.org/10.1016/j.engappai.2023.106005}}.
\newline\urlprefix\url{https://www.sciencedirect.com/science/article/pii/S0952197623001896}

\bibitem{HUSSAIN2023101}
A.~Hussain, J.~A. Memon, M.~Murshed, U.~H. Md~Shabbir~Alam, Muhammad~Rahman, A time series forecasting analysis of overall and sector-based natural gas demand: a developing south asian economy case, Environmental Science and Pollution Research 29 (2022).
\newblock \href {https://doi.org/https://doi.org/10.1007/s11356-022-20861-3} {\path{doi:https://doi.org/10.1007/s11356-022-20861-3}}.

\bibitem{en15134880}
S.-Y. Shin, H.-G. Woo, \href{https://www.mdpi.com/1996-1073/15/13/4880}{Energy consumption forecasting in korea using machine learning algorithms}, Energies 15~(13) (2022).
\newblock \href {https://doi.org/10.3390/en15134880} {\path{doi:10.3390/en15134880}}.
\newline\urlprefix\url{https://www.mdpi.com/1996-1073/15/13/4880}

\bibitem{Breiman1984ClassificationAR}
L.~Breiman, J.~H. Friedman, R.~A. Olshen, C.~J. Stone, Classification and regression trees, 1984.
\newblock \href {https://doi.org/https://doi.org/10.1201/9781315139470} {\path{doi:https://doi.org/10.1201/9781315139470}}.

\bibitem{Quinlan1986}
J.~R. Quinlan, \href{https://doi.org/10.1007/BF00116251}{Induction of decision trees}, Machine Learning 1~(1) (1986) 81--106.
\newblock \href {https://doi.org/10.1007/BF00116251} {\path{doi:10.1007/BF00116251}}.
\newline\urlprefix\url{https://doi.org/10.1007/BF00116251}

\bibitem{10.5555/583200}
J.~R. Quinlan, C4.5: Programs for Machine Learning, Morgan Kaufmann Publishers Inc., San Francisco, CA, USA, 1993.
\newblock \href {https://doi.org/https://doi.org/10.1007/BF00993309} {\path{doi:https://doi.org/10.1007/BF00993309}}.

\bibitem{10.1145/347090.347107}
P.~Domingos, G.~Hulten, \href{https://doi.org/10.1145/347090.347107}{Mining high-speed data streams}, in: Proceedings of the Sixth ACM SIGKDD International Conference on Knowledge Discovery and Data Mining, KDD '00, Association for Computing Machinery, New York, NY, USA, 2000, p. 71–80.
\newblock \href {https://doi.org/10.1145/347090.347107} {\path{doi:10.1145/347090.347107}}.
\newline\urlprefix\url{https://doi.org/10.1145/347090.347107}

\bibitem{10.1145/3219819.3220005}
C.~Manapragada, G.~I. Webb, M.~Salehi, \href{https://doi.org/10.1145/3219819.3220005}{Extremely fast decision tree}, in: Proceedings of the 24th ACM SIGKDD International Conference on Knowledge Discovery \& Data Mining, KDD '18, Association for Computing Machinery, New York, NY, USA, 2018, p. 1953–1962.
\newblock \href {https://doi.org/10.1145/3219819.3220005} {\path{doi:10.1145/3219819.3220005}}.
\newline\urlprefix\url{https://doi.org/10.1145/3219819.3220005}

\bibitem{randomforest}
H.~Abdulsalam, D.~Skillicorn, P.~Martin, Streaming random forests, 2007, pp. 225 -- 232.
\newblock \href {https://doi.org/10.1109/IDEAS.2007.4318108} {\path{doi:10.1109/IDEAS.2007.4318108}}.

\bibitem{10.1007/978-3-319-00551-5_4}
P.~Cal, M.~Wo{\'{z}}niak, Parallel hoeffding decision tree for streaming data, in: S.~Omatu, J.~Neves, J.~M.~C. Rodriguez, J.~F. Paz~Santana, S.~R. Gonzalez (Eds.), Distributed Computing and Artificial Intelligence, Springer International Publishing, Cham, 2013, pp. 27--35.
\newblock \href {https://doi.org/"10.1007/978-3-319-00551-5"} {\path{doi:"10.1007/978-3-319-00551-5"}}.

\bibitem{Lung-Yut-Fong:2015}
A.~Lung-Yut-Fong, C.~L\'evy-Leduc, O.~Capp\'e, Homogeneity and change-point detection tests for multivariate data using rank statistics, Journal de la soci\'et\'e fran\c{c}aise de statistique 156~(4) (2015) 133--162.
\newblock \href {https://doi.org/https://doi.org/10.48550/arXiv.1107.1971} {\path{doi:https://doi.org/10.48550/arXiv.1107.1971}}.

\bibitem{Truong:2020}
C.~Truong, L.~Oudre, N.~Vayatis, \href{http://arxiv.org/abs/1801.00718}{A review of change point detection methods}, CoRR abs/1801.00718 (2018).
\newblock \href {http://arxiv.org/abs/1801.00718} {\path{arXiv:1801.00718}}.
\newline\urlprefix\url{http://arxiv.org/abs/1801.00718}

\bibitem{Haynes:2014}
K.~Haynes, I.~Eckley, P.~Fearnhead, Efficient penalty search for multiple changepoint problems (12 2014).
\newblock \href {https://doi.org/https://doi.org/10.48550/arXiv.1412.3617} {\path{doi:https://doi.org/10.48550/arXiv.1412.3617}}.

\bibitem{Aminikhanghahi:2017}
S.~Aminikhanghahi, D.~J. Cook, A survey of methods for time series change point detection, Knowledge and information systems 51~(2) (2017) 339--367.
\newblock \href {https://doi.org/https://doi.org/10.1007/s10115-016-0987-z} {\path{doi:https://doi.org/10.1007/s10115-016-0987-z}}.

\bibitem{Killick:2012}
R.~Killick, P.~Fearnhead, I.~A. Eckley, \href{https://doi.org/10.1080\%2F01621459.2012.737745}{Optimal detection of changepoints with a linear computational cost}, Journal of the American Statistical Association 107~(500) (2012) 1590--1598.
\newblock \href {https://doi.org/"10.1080/01621459.2012.737745"} {\path{doi:"10.1080/01621459.2012.737745"}}.
\newline\urlprefix\url{https://doi.org/10.1080\%2F01621459.2012.737745}

\bibitem{Jackson:2005}
B.~Jackson, J.~Scargle, D.~Barnes, S.~Arabhi, A.~Alt, P.~Gioumousis, E.~Gwin, P.~Sangtrakulcharoen, L.~Tan, T.~T. Tsai, An algorithm for optimal partitioning of data on an interval, IEEE Signal Processing Letters 12~(2) (2005) 105--108.
\newblock \href {https://doi.org/10.1109/LSP.2001.838216} {\path{doi:10.1109/LSP.2001.838216}}.

\bibitem{Scott:1974}
A.~J. Scott, M.~Knott, \href{http://www.jstor.org/stable/2529204}{A cluster analysis method for grouping means in the analysis of variance}, Biometrics 30~(3) (1974) 507--512.
\newline\urlprefix\url{http://www.jstor.org/stable/2529204}

\bibitem{Adams:2007}
R.~P. Adams, D.~J.~C. Mackay, Bayesian online changepoint detection, arXiv: Machine Learning (2007).
\newblock \href {https://doi.org/https://doi.org/10.48550/arXiv.0710.3742} {\path{doi:https://doi.org/10.48550/arXiv.0710.3742}}.

\bibitem{Faber:2021}
K.~Faber, R.~Corizzo, B.~Sniezynski, M.~Baron, N.~Japkowicz, Watch: Wasserstein change point detection for high-dimensional time series data, in: 2021 IEEE International Conference on Big Data (Big Data), 2021, pp. 4450--4459.
\newblock \href {https://doi.org/10.1109/BigData52589.2021.9671962} {\path{doi:10.1109/BigData52589.2021.9671962}}.

\bibitem{SVOBODA2021119430}
R.~Svoboda, V.~Kotik, J.~Platos, \href{http://www.sciencedirect.com/science/article/pii/S0360544220325378}{Short-term natural gas consumption forecasting from long-term data collection}, Energy 218 (2021) 119430.
\newblock \href {https://doi.org/https://doi.org/10.1016/j.energy.2020.119430} {\path{doi:https://doi.org/10.1016/j.energy.2020.119430}}.
\newline\urlprefix\url{http://www.sciencedirect.com/science/article/pii/S0360544220325378}

\bibitem{TASCIKARAOGLU2014243}
A.~Tascikaraoglu, M.~Uzunoglu, \href{https://www.sciencedirect.com/science/article/pii/S1364032114001944}{A review of combined approaches for prediction of short-term wind speed and power}, Renewable and Sustainable Energy Reviews 34 (2014) 243--254.
\newblock \href {https://doi.org/https://doi.org/10.1016/j.rser.2014.03.033} {\path{doi:https://doi.org/10.1016/j.rser.2014.03.033}}.
\newline\urlprefix\url{https://www.sciencedirect.com/science/article/pii/S1364032114001944}

\bibitem{bento9854716}
P.~M. Bento, J.~A. Pombo, S.~J. Mariano, M.~R. Calado, Short-term price forecasting in the iberian electricity market: Sensitivity assessment of the exogenous variables influence (2022) 1--7\href {https://doi.org/10.1109/EEEIC/ICPSEurope54979.2022.9854716} {\path{doi:10.1109/EEEIC/ICPSEurope54979.2022.9854716}}.

\bibitem{HOSOVSKY2021101955}
A.~Hošovský, J.~Piteľ, M.~Adámek, J.~Mižáková, K.~Židek, \href{https://www.sciencedirect.com/science/article/pii/S2352710220335877}{Comparative study of week-ahead forecasting of daily gas consumption in buildings using regression arma/sarma and genetic-algorithm-optimized regression wavelet neural network models}, Journal of Building Engineering 34 (2021) 101955.
\newblock \href {https://doi.org/https://doi.org/10.1016/j.jobe.2020.101955} {\path{doi:https://doi.org/10.1016/j.jobe.2020.101955}}.
\newline\urlprefix\url{https://www.sciencedirect.com/science/article/pii/S2352710220335877}

\bibitem{woo2023deep}
G.~Woo, C.~Liu, D.~Sahoo, A.~Kumar, S.~Hoi, Learning deep time-index models for time series forecasting (2023).
\newblock \href {http://arxiv.org/abs/2207.06046} {\path{arXiv:2207.06046}}, \href {https://doi.org/https://doi.org/10.48550/arXiv.2207.06046} {\path{doi:https://doi.org/10.48550/arXiv.2207.06046}}.

\bibitem{domingos2000}
P.~Domingos, G.~Hulten, \href{https://doi.org/10.1145/347090.347107}{Mining high-speed data streams}, in: Proceedings of the Sixth ACM SIGKDD International Conference on Knowledge Discovery and Data Mining, KDD '00, Association for Computing Machinery, New York, NY, USA, 2000, p. 71–80.
\newblock \href {https://doi.org/10.1145/347090.347107} {\path{doi:10.1145/347090.347107}}.
\newline\urlprefix\url{https://doi.org/10.1145/347090.347107}

\bibitem{Hochreiter1997}
S.~Hochreiter, J.~Schmidhuber, \href{https://doi.org/10.1162/neco.1997.9.8.1735}{{Long Short-Term Memory}}, Neural Computation 9~(8) (1997) 1735--1780.
\newblock \href {http://arxiv.org/abs/https://direct.mit.edu/neco/article-pdf/9/8/1735/813796/neco.1997.9.8.1735.pdf} {\path{arXiv:https://direct.mit.edu/neco/article-pdf/9/8/1735/813796/neco.1997.9.8.1735.pdf}}, \href {https://doi.org/10.1162/neco.1997.9.8.1735} {\path{doi:10.1162/neco.1997.9.8.1735}}.
\newline\urlprefix\url{https://doi.org/10.1162/neco.1997.9.8.1735}

\bibitem{KRAWCZYK2017132}
B.~Krawczyk, L.~L. Minku, J.~Gama, J.~Stefanowski, M.~Woźniak, \href{https://www.sciencedirect.com/science/article/pii/S1566253516302329}{Ensemble learning for data stream analysis: A survey}, Information Fusion 37 (2017) 132--156.
\newblock \href {https://doi.org/https://doi.org/10.1016/j.inffus.2017.02.004} {\path{doi:https://doi.org/10.1016/j.inffus.2017.02.004}}.
\newline\urlprefix\url{https://www.sciencedirect.com/science/article/pii/S1566253516302329}

\bibitem{BifetGavaldaEtAl18}
A.~Bifet, R.~Gavald{\`a}, G.~Holmes, B.~Pfahringer, \href{https://moa.cms.waikato.ac.nz/book-html/}{Machine Learning for Data Streams with Practical Examples in MOA}, MIT Press, Cambridge, MA, 2018.
\newline\urlprefix\url{https://moa.cms.waikato.ac.nz/book-html/}

\bibitem{PETROPOULOS2022705}
F.~Petropoulos, D.~Apiletti, V.~Assimakopoulos, M.~Z. Babai, D.~K. Barrow, S.~{Ben Taieb}, C.~Bergmeir, R.~J. Bessa, J.~Bijak, J.~E. Boylan, J.~Browell, C.~Carnevale, J.~L. Castle, P.~Cirillo, M.~P. Clements, C.~Cordeiro, F.~L. {Cyrino Oliveira}, S.~{De Baets}, A.~Dokumentov, J.~Ellison, P.~Fiszeder, P.~H. Franses, D.~T. Frazier, M.~Gilliland, M.~S. Gönül, P.~Goodwin, L.~Grossi, Y.~Grushka-Cockayne, M.~Guidolin, M.~Guidolin, U.~Gunter, X.~Guo, R.~Guseo, N.~Harvey, D.~F. Hendry, R.~Hollyman, T.~Januschowski, J.~Jeon, V.~R.~R. Jose, Y.~Kang, A.~B. Koehler, S.~Kolassa, N.~Kourentzes, S.~Leva, F.~Li, K.~Litsiou, S.~Makridakis, G.~M. Martin, A.~B. Martinez, S.~Meeran, T.~Modis, K.~Nikolopoulos, D.~Önkal, A.~Paccagnini, A.~Panagiotelis, I.~Panapakidis, J.~M. Pavía, M.~Pedio, D.~J. Pedregal, P.~Pinson, P.~Ramos, D.~E. Rapach, J.~J. Reade, B.~Rostami-Tabar, M.~Rubaszek, G.~Sermpinis, H.~L. Shang, E.~Spiliotis, A.~A. Syntetos, P.~D. Talagala, T.~S. Talagala, L.~Tashman, D.~Thomakos, T.~Thorarinsdottir, E.~Todini,
  J.~R. {Trapero Arenas}, X.~Wang, R.~L. Winkler, A.~Yusupova, F.~Ziel, \href{https://www.sciencedirect.com/science/article/pii/S0169207021001758}{Forecasting: theory and practice}, International Journal of Forecasting 38~(3) (2022) 705--871.
\newblock \href {https://doi.org/https://doi.org/10.1016/j.ijforecast.2021.11.001} {\path{doi:https://doi.org/10.1016/j.ijforecast.2021.11.001}}.
\newline\urlprefix\url{https://www.sciencedirect.com/science/article/pii/S0169207021001758}

\bibitem{Diebold2002}
F.~Diebold, R.~Mariano, Comparing predictive accuracy, Journal of Business and Economic Statistics 20 (2002) 134--44.
\newblock \href {https://doi.org/10.1080/07350015.1995.10524599} {\path{doi:10.1080/07350015.1995.10524599}}.

\end{thebibliography}

\endgroup



\end{document}